\documentclass{article}

\PassOptionsToPackage{numbers,sort}{natbib}

\usepackage[preprint]{neurips_2026}

\usepackage[utf8]{inputenc}
\usepackage[T1]{fontenc}
\usepackage{hyperref}
\usepackage{url}
\usepackage{booktabs}
\usepackage{multirow}
\usepackage{placeins}     
\usepackage{amsfonts}
\usepackage{nicefrac}
\usepackage{microtype}
\usepackage{xcolor}
\usepackage{graphicx}
\usepackage{amsmath}        
\usepackage{amssymb}        
\usepackage{bm}              
\usepackage{algorithm}      
\usepackage{algpseudocode}  

\title{Spectral Unforgetting: Post-Hoc Recovery of Damaged Capabilities Without Retraining}

\author{%
  Aarash Abro \\
  Zeta Labs \\
  \texttt{aarash@zetasolutions.org} \\
  \And
  Muhammad Tahir \\
  Lahore University of Management Sciences \\
  \texttt{tahir@lums.edu.pk} \\
}

\begin{document}

\maketitle

\begin{abstract}
Fine-tuning a language model for a target task routinely degrades
capabilities the training data never explicitly threatened. We study
this phenomenon, known as catastrophic forgetting, and propose a
post-hoc repair solution that uses only the pretrained checkpoint
$W_{\mathrm{base}}$ and its fine-tuned descendant $W_{\mathrm{ft}}$.
The goal is not merely to revert the model toward the base
checkpoint, but to recover capabilities damaged by fine-tuning while
preserving both the target-task gains and any beneficial held-out
improvements. We introduce DG-Hard, a checkpoint-only spectral repair
method for the fine-tuning update
$\Delta = W_{\mathrm{ft}} - W_{\mathrm{base}}$. DG-Hard treats
$\Delta$ as a low-rank task-aligned signal embedded in an IID-like
noise residual that gradient descent has no incentive to remove, and
applies the Donoho-Gavish hard singular-value threshold to each
weight-delta matrix, keeping the structured high-energy part of the
update and removing the spectral bulk. This reduces repair to a
closed-form SVD filtering step requiring no data-dependent tuning. A
central difficulty is evaluation: average accuracy hides per-benchmark
failures, while naive recovery scores reward models that simply revert
toward the base. We therefore introduce a partition-conditional metric
that separately tracks healing, preservation, non-damage, and
target-task retention. Across $14$ (model, task) settings and nine
cross-domain held-out benchmarks, DG-Hard achieves the strongest
balanced repair among post-hoc baselines. DG-Hard also restores safety
alignment degraded by benign fine-tuning on three independent safety
axes, despite using no alignment data. These results suggest that part
of fine-tuning-induced capability loss is not an unavoidable
consequence of specialization, but a removable spectral residue in the
weight update itself.
The code can be found at:
\url{https://github.com/BrickleRex/dghard}.

\end{abstract}

%

\section{Introduction}
\label{sec:intro}

Fine-tuning can improve a model on the task it is trained for while
destroying capabilities that were already present in the pretrained
checkpoint. In our experiments, fine-tuning Qwen3.5-4B on the medical
domain raises medical-question-answering accuracy, but reduces
math-reasoning accuracy drastically. This is not an isolated failure
mode: across the $14$ (model, task) fine-tuning cells we evaluate,
$13$ exhibit at least one single-benchmark collapse
(Tab.~\ref{tab:cell-damage}, App.~\ref{app:per-cell-tables}). Such
failures are the modern large-model form of \emph{catastrophic
forgetting}~\citep{mccloskey1989catastrophic}: adaptation to a new
objective can overwrite, distort, or suppress behavior the base model
had already acquired~\citep{ratcliff1990connectionist,french1999catastrophic}.

Fine-tuning is the standard way pretrained models are
specialized~\citep{brown2020language, howard2018universal,
devlin2019bert, ouyang2022training}, yet its objective contains no
term requiring unrelated capabilities to be preserved. The
consequences are documented across general-knowledge
accuracy~\citep{luo2023empirical}, safety alignment in aligned
models~\citep{qi2024finetuning}, and the geometric distortion of
pretrained features outside the training-data
span~\citep{kumar2022finetuning}. The checkpoint moves in two ways at
once: a structured update lowering the target-task loss, and a
residual from many mini-batch SGD steps whose noise scale depends on
batch size~\citep{jastrzebski2017three, keskar2016large}. The loss
rewards only the first; the second accumulates in directions
important for other capabilities, leaving the fine-tuned checkpoint a
mixture of task-aligned signal and collateral change.

We study the post-hoc repair of forgetting that has already happened.
Given only a base checkpoint $W_{\mathrm{base}}$ and a fine-tuned
checkpoint $W_{\mathrm{ft}}$, the goal is to recover damaged held-out
capabilities while preserving the gains fine-tuning was meant to
produce, and has incidentally produced. Reversion toward the base recovers forgotten behavior
cheaply but also removes the task-aligned update; a useful repair
must distinguish the part of the fine-tuning delta
$\Delta = W_{\mathrm{ft}} - W_{\mathrm{base}}$ that carries the new
task from the part that causes collateral damage.

Existing post-hoc methods make this decision in coordinate space,
whether by scalar interpolation~(WiSE-FT,
\citealp{wortsman2022robust}), random
dropping~(DARE, \citealp{yu2023language}), magnitude- and sign-aware
pruning~(TIES, \citealp{yadav2023ties}), or forgetting-aware
pruning~(FAPM, \citealp{huang2025fapm}). All face the same
representation problem: task-relevant entries and harmful residual
entries are interleaved by magnitude, sign, and position.

The same update is more separable in singular-value space. Across
fine-tuning deltas, the spectrum splits into a broad bulk that
matches the random-matrix prediction~\citep{marchenko1967distribution}
and a smaller number of outlying spikes that carry the task-aligned
update. Both halves are independently supported in prior work:
fine-tuning weight updates are
rank-deficient~\citep{hu2022lora,aghajanyan2021intrinsic}, and
trained-weight spectra fit the Marchenko-Pastur bulk past a finite
number of outliers~\citep{thamm2022random,staats2023boundary}; we
verify both on our own deltas in App.~\ref{app:low-rank-evidence}
(formalized in \S\ref{sec:method}). Repair becomes a matrix-denoising
problem: keep the structure, revert the bulk.

We instantiate this with \textbf{DG-Hard}, a closed-form spectral
repair: each fine-tuning delta matrix's SVD is hard-thresholded at
the Donoho-Gavish cut~\citep{donoho2014optimal}, and the surviving
singular components form a spectrally pruned delta $\Delta^{*}$ that
yields the repaired checkpoint $W^{*} = W_{\mathrm{base}} + \Delta^{*}$
(Alg.~\ref{alg:dg-repair}). The method is data-free, gradient-free,
training-free, and runs in minutes on a single GPU.

\begin{figure}[t]
\centering
\includegraphics[width=\linewidth]{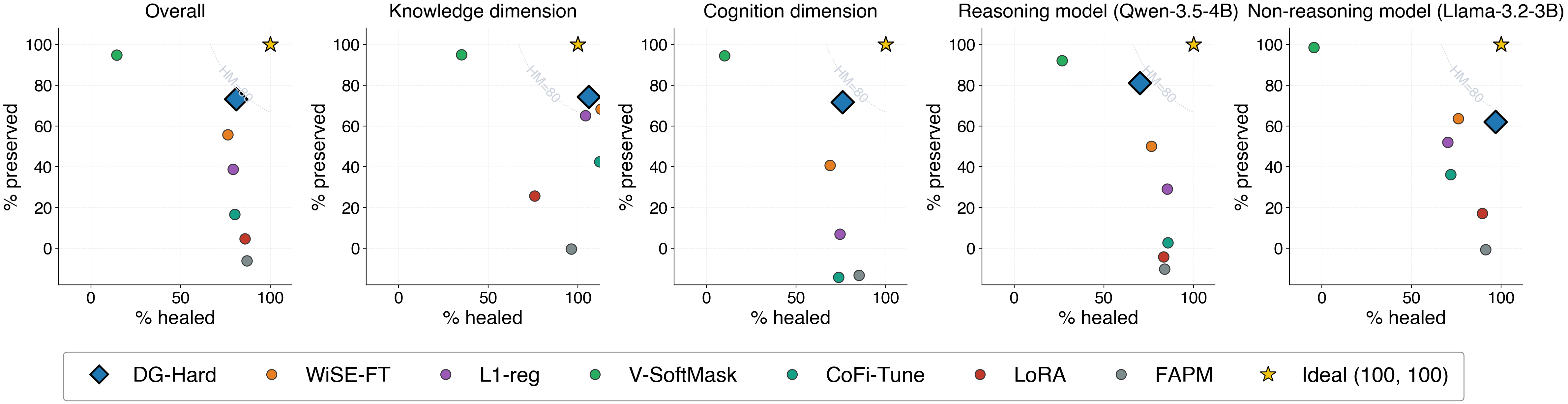}
\caption{Recovery $\times$ preservation per cohort. Each panel plots
the \% healed score on the damaged partition (x-axis) against the
\% preserved score on the improved partition (y-axis), as defined in
\S\ref{sec:recovery-preservation}. The ideal corner is $(100,100)$,
and the dotted contour marks
$\mathrm{HM}(\text{\% healed},\,\text{\% preserved}) = 80$.
DG-Hard (blue diamond) is closest to the ideal corner across all five
cohorts. FAPM~\citep{huang2025fapm} strongly recovers damaged
measurements but sacrifices improved ones;
V-SoftMask~\citep{ke2023continual} preserves improved measurements
but recovers less damage.}
\label{fig:recovery-preservation-scatter}
\end{figure}

\paragraph{Contributions.}
\begin{enumerate}
\item We formulate post-hoc repair of catastrophic forgetting as a
recovery-preservation problem: recover damaged held-out capabilities
from a fine-tuned checkpoint while preserving target-task gains and
incidental held-out improvements.

\item We identify a spectral structure in fine-tuning deltas.
Empirically, damaging residuals concentrate in the singular-value
bulk, while task-relevant updates appear as singular-value spikes
(App.~\ref{app:low-rank-evidence}). This explains why scalar
interpolation and coordinate-wise pruning face an unfavorable
trade-off. We propose \textbf{DG-Hard}, a data-free repair method
that applies the Donoho-Gavish hard threshold~\citep{donoho2014optimal}
to each delta matrix and reconstructs the checkpoint from retained
singular components.

\item We introduce a partition-conditional evaluation that reports
recovery on damaged measurements, preservation on improved
measurements, and retention on unchanged and target-task measurements.
DG-Hard achieves the best recovery-preservation trade-off among
post-hoc baselines across $14$ (model, task) cells and $9$ held-out
benchmarks (Tab.~\ref{tab:headline-method},
Fig.~\ref{fig:recovery-preservation-scatter}).
\end{enumerate}

%

\section{Background and related work}
\label{sec:related-work}

\paragraph{Catastrophic forgetting.}
We extend the connectionist account of
\citet{mccloskey1989catastrophic} that distributed representations
make any helpful update perturb the weights encoding prior tasks
\citep{ratcliff1990connectionist, french1999catastrophic}.
\citet{kumar2022finetuning} formalize the modern instance in an
overparameterized linear setting, and \citet{luo2023empirical,
qi2024finetuning} document its empirical reproduction in contemporary
LLMs across general-knowledge and safety benchmarks.

\paragraph{Training-time approaches.}
Training-time methods modify the optimization step itself and fall
into three subfamilies: parameter-movement regularizers,
gradient-masking via per-unit importance, and replay against stored
past-task data. Parameter-movement regularizers penalize updates
weighted by per-parameter importance and differ mostly in how
importance is estimated: \textbf{EWC}~\citep{kirkpatrick2017overcoming}
uses the Fisher diagonal, \textbf{SI}~\citep{zenke2017continual} a
path-integral measure of online contribution to the previous-task
loss, and \textbf{MAS}~\citep{aljundi2018memory} the gradient of the
squared L2 norm of the network output (label-free, since it
estimates importance from unlabeled calibration data). We additionally
compare against a generic \textbf{L1-reg} baseline that penalizes
$\|W - W_{\mathrm{base}}\|_1$ without task information. Gradient-masking
methods identify which units carry pretrained capability and slow the
gradient flow through them: \textbf{DAS}~\citep{ke2023continual}
multiplies each gradient by $(1 - \text{importance})$ where
importance comes from a dropout-KL proxy on calibration data, and
\textbf{CoFiTune}~\citep{zhang2024cofitune} extends it with a
coarse-grained layer-range filter (restricting the procedure to an
empirically selected slice of layers) and a KL/dropout-robustness
fine score. Replay-based methods mix or project against stored
past-task examples (\textbf{GEM}~\citep{lopez2017gradient},
\textbf{Experience Replay}~\citep{chaudhry2019tiny}) but assume
access to a representative pretraining sample, which is unavailable
for modern LLMs whose pretraining corpora are proprietary and at
terabyte scale.

\paragraph{Parameter-efficient fine-tuning.}
\textbf{LoRA}~\citep{hu2022lora} constrains the fine-tuning update
on each weight matrix to a low-rank factorization $BA$ with
$r \ll \min(m,n)$, leaving the base weights frozen; the deployed
model still merges base + LoRA at inference, so forgetting can
persist in the merged weights. \citet{shuttleworth2025lora} show via
direct spectral comparison that this low-rank constraint forces
LoRA-FT updates to introduce ``intruder'' singular directions
approximately orthogonal to the pretrained spectrum, and that these
intruder directions causally drive forgetting (verified by post-hoc
intervention on their singular values).

\paragraph{Post-hoc model merging.}
The family closest to our work operates post-hoc on
$\Delta = W_{\mathrm{ft}} - W_{\mathrm{base}}$ via cheap, data-free
transformations. These methods originally targeted multi-task model
merging, where multiple fine-tuned checkpoints are combined into
one; with a single fine-tune their multi-task aggregation steps
(sign election, mean over task vectors) reduce to identity, leaving
the per-vector preprocessing as a post-hoc repair on the
$(W_{\mathrm{base}}, W_{\mathrm{ft}})$ pair.
\textbf{WiSE-FT}~\citep{wortsman2022robust} interpolates linearly
between the checkpoints, $W^{*} = (1{-}\alpha) W_{\mathrm{base}} +
\alpha W_{\mathrm{ft}}$, trading pretrained and fine-tuned behavior
with a single scalar.
\textbf{Task Arithmetic}~\citep{ilharco2023editing} introduces
$\Delta$ as a task vector and studies algebraic operations on it
(negation, addition, analogies) across multiple tasks.
\textbf{TIES-Merging}~\citep{yadav2023ties} trims low-magnitude
entries of each task vector, elects a per-parameter consensus sign
across vectors, and averages only the entries aligned with that
sign, addressing redundancy and sign disagreement as sources of
merge interference.
\textbf{DARE}~\citep{yu2023language} randomly drops $\Delta$ entries
with probability $p$ and rescales survivors by $1/(1{-}p)$, designed
as a preprocessing step that sparsifies fine-tuning deltas before
merging.
\textbf{FAPM}~\citep{huang2025fapm} scores each entry with a
forgetting-aware criterion combining absolute change magnitude and a
relative-change penalty against the pretrained weight, then prunes
low-scoring entries.
\textbf{Fisher-weighted averaging}~\citep{matena2022merging} weights
each parameter by its empirical Fisher importance, additionally
requiring a small labeled training sample. Per-method
hyperparameters are in App.~\ref{app:method-configs}.

\paragraph{Spectral shrinkage and our distinction.}
The spectral framing fits fine-tuning because the weight update is
empirically rank-deficient (\citealp{hu2022lora}; broader evidence
in App.~\ref{app:low-rank-evidence}), and random-matrix theory
supplies the matching denoiser. The Marchenko-Pastur
theorem~\citep{marchenko1967distribution} fixes the asymptotic
spectrum of an IID-noise matrix, and the Donoho-Gavish hard
threshold~\citep{donoho2014optimal} identifies the AMSE-optimal cut
between signal and noise (\citealp{gavish2017optimal} extend to
optimal continuous shrinkers). Where prior post-hoc methods operate
in coordinate space (TIES, DARE, FAPM), as a single scalar (WiSE-FT,
Task Arithmetic), or via labeled-data Fisher weighting, we operate
in singular-value space and apply this statistically optimal hard
threshold, to our knowledge the first such application to
fine-tuning weight deltas for post-hoc forgetting repair.


%

\providecommand{\Wft}{W_{\mathrm{ft}}}
\providecommand{\Wbase}{W_{\mathrm{base}}}
\providecommand{\Wstar}{W^{*}}
\providecommand{\Dstar}{\Delta^{*}}
\providecommand{\sighat}{\hat{\sigma}}
\providecommand{\lammp}{\lambda_{\text{MP}}}
\providecommand{\diag}{\operatorname{diag}}
\providecommand{\R}{\mathbb{R}}
\providecommand{\dghard}{\eta_{\mathrm{DG}}}
\providecommand{\med}{\operatorname{median}}

\section{Method}
\label{sec:method}

We propose \textbf{DG-Hard}: a post-hoc, data-free, gradient-free
repair that strips the IID noise residual from a fine-tuning update
by applying the optimal hard singular-value threshold of
\citet{donoho2014optimal} to its SVD. The procedure takes only the
base checkpoint $\Wbase$ and the fine-tuned checkpoint $\Wft$,
operates one $2$D weight matrix at a time, and reduces to a
closed-form threshold in the matrix shape and a single estimated
noise scale.

\subsection{Signal-plus-noise model of the fine-tuning update}
\label{sec:setup}

For a single $2$D weight matrix in $\R^{m \times n}$, write the
fine-tuning update as $\Delta = \Wft - \Wbase$ and take its singular
value decomposition,
\begin{equation}
\label{eq:svd}
\Delta \;=\; U\,\diag(\mathbf{s})\,V^{\!\top}, \quad s_1 \geq \cdots \geq s_p \geq 0, \quad p = \min(m, n).
\end{equation}
We model $\Delta$ as a two-component sum,
\begin{equation}
\label{eq:s+n}
\Delta \;=\; \Delta_{\mathrm{signal}} + \Delta_{\mathrm{noise}},
\end{equation}
with $\Delta_{\mathrm{signal}}$ a low-rank task-aligned update and
$\Delta_{\mathrm{noise}}$ an IID residual of unknown scale $\sigma$.
The SGD-noise mechanism in \S\ref{sec:intro} produces exactly this
structure: thousands of mini-batch residuals accumulate into the
weights while the gradient has no incentive to remove them.
Both components are empirically supported: the fine-tuning weight
update is rank-deficient \citep{hu2022lora} and fine-tuning more
broadly has low intrinsic dimensionality
\citep{aghajanyan2021intrinsic}; trained-weight spectra fit the
Marchenko-Pastur bulk past a finite number of outliers
\citep{thamm2022random, staats2023boundary}. A direct check on
our own deltas confirms both properties layer-locally
(App.~\ref{app:low-rank-evidence}, Fig.~\ref{fig:spectral-unforgetting}).
Our goal is to remove $\Delta_{\mathrm{noise}}$ while leaving
$\Delta_{\mathrm{signal}}$ intact.

\subsection{Random-matrix theory and the bulk edge}
\label{sec:rmt}

If $\Delta$ were pure IID noise of variance $\sigma^{2}$, the
Marchenko-Pastur theorem~\citep{marchenko1967distribution} pins its
spectrum exactly: in the limit $m, n \to \infty$ at fixed aspect ratio
$\beta = \min(m,n)/\max(m,n)$, every singular value sits below the bulk edge
\begin{equation}
\label{eq:mp-edge}
\lammp \;=\; \sigma\,\bigl(1 + \sqrt{\beta}\bigr)
\end{equation}
with probability one, depending only on $\sigma$ and $\beta$.
Contrapositively, any $s_i > \lammp$ in $\Delta$ must reflect the
signal component of \eqref{eq:s+n}. The edge is a principled boundary
but not a denoising rule; for that we need a \emph{shrinker} mapping
each $s_i$ to a denoised $s'_i$. The simplest principled choice is a
hard threshold, and \citet{donoho2014optimal} derive the AMSE-optimal
one. Asymptotic mean squared error (AMSE) is the $m, n \to \infty$
limit of the expected reconstruction error
$\mathbb{E}\bigl[\|\hat{\Delta}_{\mathrm{signal}} -
\Delta_{\mathrm{signal}}\|_F^2\bigr]$ (squared Frobenius distance
between the estimated and true low-rank signal) at fixed aspect
ratio $\beta$.
A low threshold admits noise (high variance); a high threshold
discards signal (high bias); the AMSE-optimal threshold sits at the
trade-off minimum.

\subsection{DG hard: the optimal hard threshold}
\label{sec:dghard}

Among all hard shrinkers $\eta(s) = s \cdot \mathbf{1}\{s > \tau\}$,
\citet{donoho2014optimal} identify the threshold minimizing the
AMSE of the reconstruction under \eqref{eq:s+n}:
\begin{equation}
\label{eq:dg-hard-tau}
\tau^{*} = \omega(\beta)\,\sigma\,\sqrt{\max(m, n)}, \qquad \omega(\beta) = \sqrt{\,2(\beta+1) + \tfrac{8\beta}{(\beta+1) + \sqrt{\beta^{2} + 14\beta + 1}}\,}.
\end{equation}
For square matrices, $\omega(1) = 4/\sqrt{3} \approx 2.309$, the
constant for which the original paper is named. The DG-Hard shrinker is
\begin{equation}
\label{eq:dg-hard}
\dghard(s) \;=\; s \cdot \mathbf{1}\!\bigl\{\,s > \tau^{*}\bigr\}.
\end{equation}
$\tau^{*}$ sits strictly above $\lammp$ ($\approx 2.309\,\sigma\sqrt{n}$
vs.\ $2\sigma\sqrt{n}$ at $\beta = 1$): cutting at the edge would admit
singular values barely distinguishable from noise, paying full
reconstruction variance for negligible signal. Eq.~\eqref{eq:dg-hard-tau}
is the unique threshold at which marginal bias and variance balance.
Algorithm~\ref{alg:dg-repair} gives the full per-matrix procedure. The
noise estimator $\sighat$ on line~\ref{algline:sighat} is the
median-based Donoho-Gavish estimator
$\sighat = \med(\mathbf{s})/(\mu_\beta\sqrt{\max(m,n)})$, which
substitutes into \eqref{eq:dg-hard-tau} to give a data-only threshold
$\tau^{*} = (\omega(\beta)/\mu_\beta)\,\med(\mathbf{s}) \approx
2.858\,\med(\mathbf{s})$ for square matrices; full derivation and the
SVD-dominated cost ($O(\min(m,n)^{2}\max(m,n))$ per matrix) are in
App.~\ref{app:noise-scale}.

\begin{algorithm}[t]
\caption{DG hard repair (per weight matrix).}
\label{alg:dg-repair}
\begin{algorithmic}[1]
\Require Base weight $\Wbase \in \R^{m \times n}$,
         fine-tuned weight $\Wft \in \R^{m \times n}$.
\Ensure  Repaired weight $\Wstar \in \R^{m \times n}$.
\State $\Delta \gets \Wft - \Wbase$
\State $(U, \mathbf{s}, V) \gets \mathrm{SVD}(\Delta)$
\State $p \gets \min(m, n);\;\; \beta \gets p / \max(m, n)$
\State $\sighat \gets \med(\mathbf{s}) /
       \bigl(\mu_{\beta}\sqrt{\max(m, n)}\bigr)$
       \label{algline:sighat}
       \Comment{noise scale, App.~\ref{app:noise-scale}}
\State $\tau \gets \omega(\beta)\,\sighat\,\sqrt{\max(m,n)}$
       \Comment{optimal threshold, \eqref{eq:dg-hard-tau}}
\State $s'_i \gets s_i \cdot \mathbf{1}\{s_i > \tau\}$
       \textbf{ for } $i = 1, \ldots, p$
\State $\Dstar \gets U\,\diag(\mathbf{s}')\,V^{\!\top}$
\State \Return $\Wbase + \Dstar$
\end{algorithmic}
\end{algorithm}

\subsection{Evaluation protocol}
\label{sec:setup-eval}

For each held-out benchmark $b$ in every (model, task, method) cell
we read $\mathrm{score}_b \in [0, 1]$ and define
\begin{equation}
\label{eq:delta-defs}
\Delta_{\mathrm{FT}}(b) = \mathrm{score}_{\mathrm{FT}}(b) - \mathrm{score}_{\mathrm{base}}(b), \qquad \Delta_{\mathrm{method}}(b) = \mathrm{score}_{\mathrm{method}}(b) - \mathrm{score}_{\mathrm{base}}(b).
\end{equation}
At a $\pm 3$\,pp threshold each (model, task, $b$) triple is
\textbf{damaged} if $\Delta_{\mathrm{FT}}(b) \leq -3$,
\textbf{improved} if $\Delta_{\mathrm{FT}}(b) \geq +3$, and
\textbf{unchanged} otherwise. The partition is fixed by the FT
checkpoint and identical across methods, so per-method comparisons
operate on the same set of triples in each partition cell. Across the $126$-cell matrix ($2$ models
$\times$ $7$ tasks $\times$ $9$ benchmarks) we observe $30$ damaged,
$55$ improved, $41$ unchanged triples.

\paragraph{Per-method statistics.}
Let $s_M = \mathrm{score}_{\mathrm{method}}$,
$s_F = \mathrm{score}_{\mathrm{FT}}$, and
$s_B = \mathrm{score}_{\mathrm{base}}$. Let $D, I, U$ denote the
damaged, improved, and unchanged partition sets, and HM the harmonic
mean. Each method produces seven percentage scores. Higher is
better; $100$ marks full recovery or preservation, values above
$100$ indicate overshoot, values below $0$ indicate regression, and
HM aggregates are floored at $0$:
\begin{subequations}
\label{eq:metric-defs}
\begin{align}
\textbf{\% healed}
  &= \mathop{\mathrm{avg}}_{b\in D}\;\dfrac{s_M(b) - s_F(b)}{s_B(b) - s_F(b)}\cdot 100,
\quad
\textbf{\% preserved} = \mathop{\mathrm{avg}}_{b\in I}\;\dfrac{s_M(b) - s_B(b)}{s_F(b) - s_B(b)}\cdot 100, \\[6pt]
\textbf{on-task ret.}
  &= \mathop{\mathrm{avg}}_{\text{cells}}\;\dfrac{s_M(\mathrm{task})}{s_F(\mathrm{task})}\cdot 100,
\quad
\textbf{non-damage} = \mathop{\mathrm{avg}}_{b\in U}\,\mathbf{1}\{s_B(b) - s_M(b) < 3\}\cdot 100, \\[6pt]
\textbf{Clean-up}
  &= \mathrm{HM}(\text{\% healed},\, \text{non-damage}),
\quad
\textbf{Retention} = \mathrm{HM}(\text{\% preserved},\, \text{on-task ret.}),
\end{align}
\begin{equation}
\label{eq:combined-def}
\textbf{Combined} = \mathrm{HM}(\text{Clean-up},\, \text{Retention}).
\end{equation}
\end{subequations}
Each statistic measures one axis: \textbf{\% healed} (FT damage
recovered on $D$), \textbf{\% preserved} (FT gain retained on $I$),
\textbf{on-task ret.} (on-task accuracy held vs.\ FT),
\textbf{non-damage} (unchanged triples kept within $3$\,pp of base),
and \textbf{Clean-up}/\textbf{Retention} (HM of the corresponding pair).
``Combined'' is the ranking statistic and acts as a
triple-bottleneck harmonic mean: collapse on any sub-statistic drives
it toward $0$, demoting methods that win one axis by sacrificing
another (V-SoftMask preserves but does not heal; FAPM heals but does
not preserve). Per-cell scores substitute a mean-ratio variant; both
aggregations and a comparison to flat unified
scores~\citep{zhang2024cofitune} are in
App.~\ref{app:metric-discussion}.


%

\section{Experiments}
\label{sec:experiments}

A \emph{cell} is one (model, task, held-out benchmark) triple, and a
\emph{(model, task) cell} is one of the $14$ fine-tuned checkpoints,
each expanding into $9$ benchmark cells for a $126$-cell matrix. The
headline finding across this matrix: DG-Hard sits at the highest
balanced point of the recovery-preservation trade-off, dominating on
the reasoning model and matching the strongest baseline on the
non-reasoning model. Full per-method per-cell scores are in
App.~\ref{app:percell-bench}.

\subsection{Setup at a glance}
\label{sec:setup-methods}

Two models span the reasoning / non-reasoning split:
\textbf{Qwen3.5-4B} (run in thinking mode with \texttt{<think>}
blocks stripped at scoring) and \textbf{Llama-3.2-3B-Instruct}. Each
is fine-tuned under a uniform full-parameter SFT configuration on
seven tasks (RTE, StrategyQA, ReClor, BoolQ, MedQA, WikiQA,
Winogrande) and evaluated on nine cross-domain held-out benchmarks,
partitioned into a 3-benchmark \emph{Knowledge} cohort (MMLU,
TriviaQA, TruthfulQA) and a 6-benchmark \emph{Cognition} cohort
(ARC-Challenge, GSM8K, IFEval, Math-500, MNLI, HellaSwag). The nine
compared methods fall into three groups: reference points
\textbf{Base} (post-alignment, unfine-tuned) and \textbf{Full-SFT}
(unrepaired fine-tuned); training-time interventions \textbf{L1-reg},
\textbf{V-SoftMask} (DAS,~\citealp{ke2023continual}),
\textbf{CoFi-Tune}~\citep{zhang2024cofitune}, and
\textbf{LoRA}~\citep{hu2022lora} (rank $16$); and post-hoc methods
\textbf{WiSE-FT}~\citep{wortsman2022robust},
\textbf{FAPM}~\citep{huang2025fapm}, and our \textbf{DG-Hard}, the
only spectral method in the set; element-wise merging baselines
(TIES~\citep{yadav2023ties}, DARE~\citep{yu2023language},
Task Arithmetic~\citep{ilharco2023editing}) appear in
App.~\ref{app:abl-merging}. Inference is identical across methods (vLLM,
greedy decoding, no \texttt{max\_tokens} cap), with a measured $\pm
1$\,pp drift floor and a $3$\,pp significance threshold
($\approx 3\times$ the drift floor) applied throughout. HuggingFace
identifiers, eval splits, hyperparameters, per-model repair scope,
method configurations, and inference details are in
App.~\ref{app:setup}, \ref{app:method-configs}, and~\ref{app:inference}.

\subsection{Per-cell forgetting and repair winners}
\label{sec:per-cell-forgetting}
\label{sec:per-cell-balance}

Fine-tuning damages at least one held-out benchmark in $13$ of $14$
cells, often catastrophically (Qwen/MedQA's GSM8K collapses from
$93.0\%$ to $1.1\%$). Damage concentrates on a small number of
benchmarks per cell rather than spreading uniformly, and the matrix
contains $55$ improved triples against only $30$ damaged ones, so a
cell-level mean held-out score smooths the catastrophic drops away
and is the wrong instrument to detect forgetting. The rest of the
section therefore evaluates methods at the per-benchmark level
(per-cell breakdown in Tab.~\ref{tab:cell-damage},
App.~\ref{app:per-cell-tables}).

A per-cell balance score, $\mathrm{HM}$ of the method's mean
held-out ratio against base and its on-task ratio against Full-SFT
($\times 100$), resolves the per-cell winners.
\textbf{DG-Hard wins $6$ of $7$ Qwen cells (plus $1$ tie); WiSE-FT
wins $4$ of $7$ Llama cells (plus $2$ ties); on the $14$-cell union,
DG-Hard $7$, WiSE-FT $4$, ties $3$.} The reasoning-trained model is
more sensitive to the Cognition-side preservation gap that DG-Hard's
spectral threshold widens against the linear average
(\S\ref{sec:knowledge-cognition}); per-cell scores and full
discussion in Tab.~\ref{tab:cell-balance}
(App.~\ref{app:per-cell-tables}).

\subsection{Recovery vs preservation (population-level)}
\label{sec:recovery-preservation}

Recovery and preservation are coupled: reverting
$W_{\mathrm{ft}}$ toward $W_{\mathrm{base}}$ trivially
maximizes recovery but discards the held-out and on-task gains
fine-tuning produced. Tab.~\ref{tab:headline-method} reports the
population-level Clean-up, Retention, and Combined scores
(Eq.~\ref{eq:metric-defs}, \S\ref{sec:setup-eval}) for every
method.

\begin{table}[t]
\centering
\caption{Population-level Clean-up, Retention, and Combined scores
per method. DG-Hard is the only method scoring $\geq 80$ on both
Clean-up and Retention; WiSE-FT~\citep{wortsman2022robust} comes closest but falls to $71.0$ on
Retention. Every other baseline collapses on at least one axis.}
\label{tab:headline-method}
\small
\setlength{\tabcolsep}{4pt}
\resizebox{0.85\linewidth}{!}{%
\begin{tabular}{lrrrrrrr}
\toprule
Method & \% healed $\uparrow$ & Non-damage $\uparrow$ & \% preserved $\uparrow$ & On-task ret. $\uparrow$ & Clean-up $\uparrow$ & Retention $\uparrow$ & Combined $\uparrow$ \\
\midrule
\textbf{DG-Hard} & $80.9_{\pm 5.0}$ & $85.4_{\pm 2.0}$ & $73.1_{\pm 2.5}$ & $97.3_{\pm 0.2}$ & $83.1_{\pm 2.8}$ & $83.5_{\pm 1.6}$ & \textbf{83.3} \\
WiSE-FT          & $76.3_{\pm 4.4}$ & $90.2_{\pm 1.7}$ & $55.8_{\pm 2.7}$ & $97.5_{\pm 0.2}$ & $82.7_{\pm 2.7}$ & $71.0_{\pm 2.2}$ & 76.4 \\
L1-reg           & $79.3_{\pm 4.1}$ & $87.8_{\pm 2.0}$ & $38.7_{\pm 3.3}$ & $98.6_{\pm 0.2}$ & $83.4_{\pm 2.6}$ & $55.6_{\pm 3.5}$ & 66.7 \\
CoFi-Tune        & $80.2_{\pm 4.8}$ & $75.6_{\pm 2.2}$ & $16.7_{\pm 4.3}$ & $96.2_{\pm 0.2}$ & $77.8_{\pm 2.7}$ & $28.5_{\pm 6.2}$ & 41.7 \\
V-SoftMask       & $14.3_{\pm 4.5}$ & $\bm{97.6_{\pm 1.9}}$ & $\bm{94.7_{\pm 2.3}}$ & $\bm{99.6_{\pm 0.2}}$ & $24.9_{\pm 5.9}$ & $\bm{97.1_{\pm 1.2}}$ & 39.7 \\
LoRA             & $85.8_{\pm 4.8}$ & $85.4_{\pm 2.0}$ & $4.7_{\pm 3.3}$  & $94.9_{\pm 0.2}$ & $85.6_{\pm 2.7}$ & $9.0_{\pm 3.5}$  & 16.3 \\
FAPM             & $\bm{86.9_{\pm 4.4}}$ & $97.6_{\pm 1.7}$ & $-6.2_{\pm 3.9}$ & $77.8_{\pm 0.2}$ & $\bm{91.9_{\pm 2.7}}$ & $0.0_{\pm 0.3}$ & 0.0 \\
\bottomrule
\end{tabular}%
}
\end{table}

\paragraph{Per-method profiles.}
\textbf{DG-Hard} is the only method scoring $\geq 80$ on both axes
(Combined $83.3$). \textbf{FAPM} ($91.9 / 0.0$) and
\textbf{V-SoftMask} ($24.9 / 97.1$) are the one-axis extremes:
FAPM's $90\%$ sparsity reversion drags improved benchmarks $-6.2\%$
below base, while V-SoftMask's gradient damping preserves the FT
distribution but does not actively heal. \textbf{LoRA} ($85.6 / 9.0$)
cleans up well but has no FT-side gain to preserve because low-rank
adapters never produced FT's incidental held-out gains. CoFi-Tune, L1-reg, and
WiSE-FT are middling on both.

\subsection{Cohort breakdown}
\label{sec:two-views}

\begin{figure}[t]
\centering
\includegraphics[width=\linewidth]{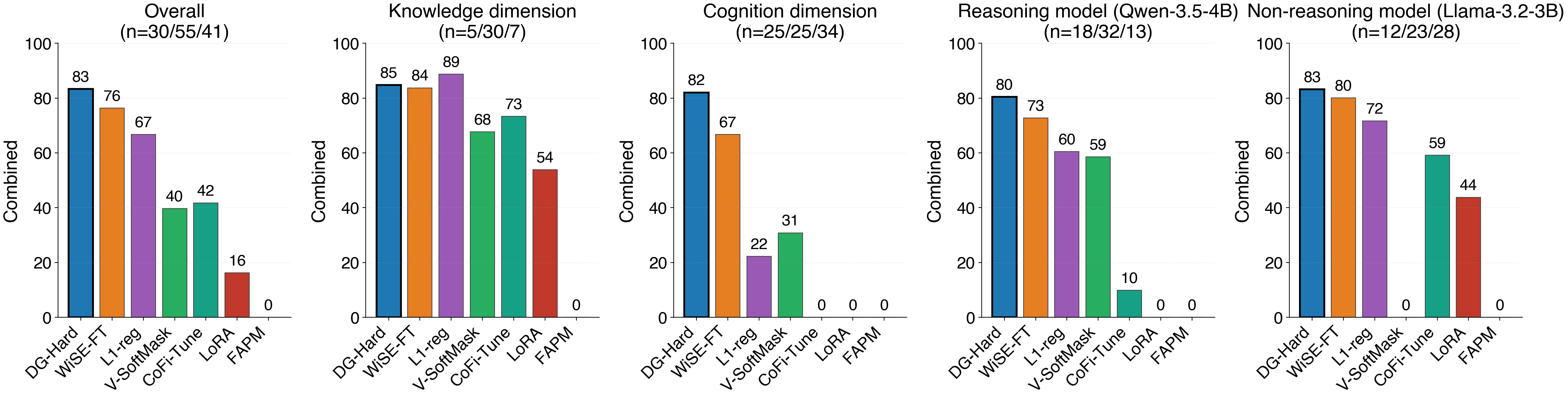}
\caption{Population-level Combined score per method, sliced by cohort.
Panel titles list $n = (\text{damaged}/\text{improved}/\text{unchanged})$
triple counts per cohort. DG-Hard tops Overall, Cognition, Reasoning,
and Non-reasoning; L1-reg edges past DG-Hard on the small-$n$
Knowledge cohort, where its $102.0$ Clean-up reflects \% healed
overshooting base on the $5$-case damaged partition. The per-cell
balance view in Tab.~\ref{tab:cell-balance} resolves the Llama cohort
into split wins between DG-Hard and WiSE-FT~\citep{wortsman2022robust}. Methods that collapse
on either Clean-up or Retention drop to a low Combined via the
harmonic mean's bottlenecking property.}
\label{fig:combined-by-cohort}
\end{figure}

\textbf{DG-Hard wins Combined on four of five cohorts} (Overall,
Cognition, Reasoning, Non-reasoning) by maintaining strong Clean-up
and strong Retention simultaneously (Fig.~\ref{fig:combined-by-cohort};
full per-cohort Clean-up and Retention in
App.~\ref{app:cohort-cleanup-retention}). On the small-$n$ Knowledge
cohort ($5$ damaged triples) L1-reg overshoots base in healing and
edges past on Combined; this advantage does not transfer to the
larger Cognition cohort, where L1-reg's Retention collapses to
$13.0$ and its Combined falls to $22.3$.
\textbf{DG-Hard's edge over WiSE-FT comes mostly from Retention}:
Clean-up scores are within $\sim 3$\,pp on most cohorts, but
Retention is $12{+}$\,pp higher on Overall, Cognition, and Reasoning,
driven by held-out preservation since on-task retention is
essentially flat at $\sim 97\%$ for both.

\subsection{Bucketed view}
\label{sec:damage-buckets}

\begin{table}[t]
\centering
\footnotesize
\begin{minipage}[t]{0.49\linewidth}
\centering
\caption{Improvement-bucket \% preserved (\%, higher is better).
Buckets sum to the $55$ (model, task, benchmark) triples in the
$126$-cell matrix.}
\label{tab:improvement-buckets}
\setlength{\tabcolsep}{3pt}
\begin{tabular*}{\linewidth}{@{\extracolsep{\fill}}lcrr@{}}
\toprule
Bucket & $n$ & DG-Hard & WiSE-FT \\
\midrule
3-10 (mild)      & 25 & $\bm{+65.8_{\pm 5.4}}$ & $+45.3_{\pm 5.8}$ \\
10-20 (moderate) & 18 & $\bm{+78.4_{\pm 0.9}}$ & $+70.6_{\pm 0.9}$ \\
20-40 (large)    & 12 & $\bm{+80.4_{\pm 0.5}}$ & $+55.4_{\pm 0.6}$ \\
\multicolumn{4}{c}{\rule{0pt}{2.4ex}} \\
\bottomrule
\end{tabular*}
\end{minipage}\hfill
\begin{minipage}[t]{0.49\linewidth}
\centering
\caption{Damage-bucket \% healed (\%, higher is better). Buckets
sum to the $30$ damaged (model, task, benchmark) triples in the
$126$-cell matrix.}
\label{tab:damage-buckets}
\setlength{\tabcolsep}{3pt}
\begin{tabular*}{\linewidth}{@{\extracolsep{\fill}}lcrr@{}}
\toprule
Bucket & $n$ & DG-Hard & WiSE-FT \\
\midrule
3-10 (mild)      & 19 & $\bm{+74.6_{\pm 7.0}}$ & $+65.1_{\pm 6.3}$ \\
10-20 (moderate) &  7 & $+94.6_{\pm 1.7}$      & $\bm{+95.1_{\pm 1.8}}$ \\
20-40 (severe)   &  3 & $+86.3_{\pm 1.7}$      & $\bm{+98.1_{\pm 1.7}}$ \\
40$+$ (extreme)  &  1 & $+87.6_{\pm 0.4}$      & $\bm{+92.3_{\pm 0.4}}$ \\
\bottomrule
\end{tabular*}
\end{minipage}
\end{table}

\begin{figure}[t]
\centering
\begin{minipage}[b]{0.48\linewidth}
\centering
\includegraphics[width=\linewidth]{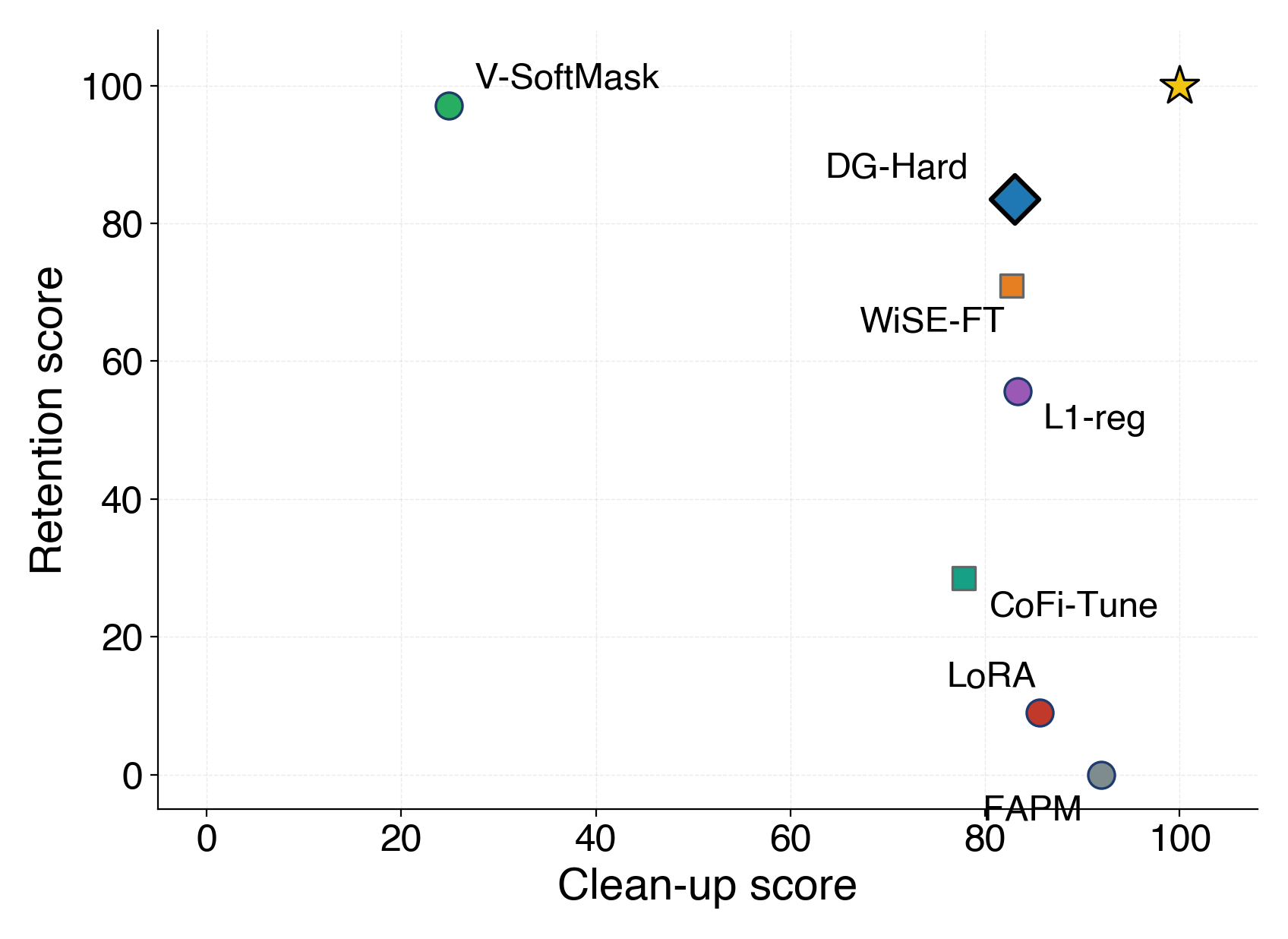}
\end{minipage}\hfill
\begin{minipage}[b]{0.48\linewidth}
\centering
\includegraphics[width=\linewidth]{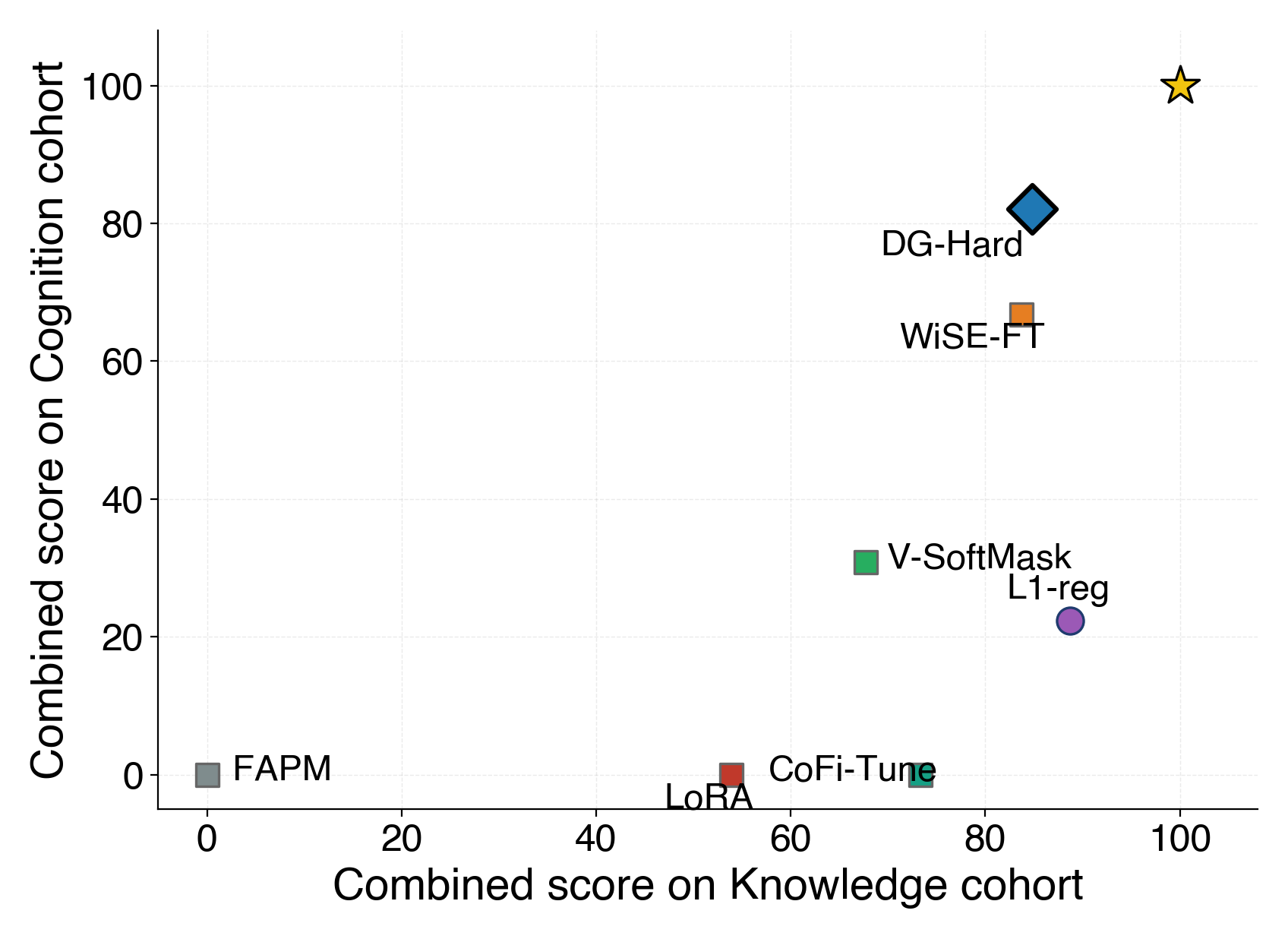}
\end{minipage}
\caption{Trade-off frontiers (both axes $0$ to $100$, higher is
better). \emph{Left}: Clean-up vs Retention. DG-Hard sits in the
upper-right region where both axes are simultaneously high;
V-SoftMask~\citep{ke2023continual} is the retention-extreme (top-left); FAPM~\citep{huang2025fapm} is the
cleanup-extreme (bottom-right). \emph{Right}: Knowledge-cohort
Combined ($x$) vs Cognition-cohort Combined ($y$). DG-Hard sits high
on both ($84.8$ and $82.1$), the most balanced strong method; WiSE-FT~\citep{wortsman2022robust}
matches on Knowledge but loses ground on Cognition ($66.8$) because
its preservation drops there; V-SoftMask sits in the
Knowledge-favouring region ($67.7$ vs $30.9$); CoFi-Tune~\citep{zhang2024cofitune}, FAPM, and LoRA~\citep{hu2022lora} collapse
on Cognition due to non-positive average preservation.}
\label{fig:pareto-cleanup-retention}
\label{fig:pareto-knowledge-cognition}
\end{figure}

Bucketing damaged and improved triples by FT-effect magnitude
(Tabs.~\ref{tab:improvement-buckets}, \ref{tab:damage-buckets};
per-cohort sub-score breakdown in App.~\ref{app:cohort-full})
resolves where the DG-Hard / WiSE-FT trade-off is actually decided.

\paragraph{Damage-recovery is roughly tied.}
Both methods heal most of the damage in every bucket
(Tab.~\ref{tab:damage-buckets}); the trade-off is not decided here.

\paragraph{Preservation widens with FT-gain magnitude.}
The preservation gap grows with the size of the FT lift
(Tab.~\ref{tab:improvement-buckets}): on the largest lifts WiSE-FT
loses about half the gain while DG-Hard keeps four-fifths. This
matches the spectral mechanism: DG-Hard's threshold retains the
high-singular-value directions of $\Delta$ that a linear weight
average uniformly attenuates, so the more concentrated the FT
signal, the wider DG-Hard's preservation lead. The damage side runs
the same direction in expectation: the mild bucket holds $19$ of the
$30$ damaged triples and DG-Hard wins it, so even where average
healing reads as a tie the dominant mode of forgetting falls to
DG-Hard (Tab.~\ref{tab:damage-buckets}).

\subsection{Cognition and the trade-off frontier}
\label{sec:knowledge-cognition}
\label{sec:pareto}

The Knowledge cohort ($n = 5$ damaged) is too easy to differentiate
methods on healing: every method except V-SoftMask recovers
at or above base on the damaged Knowledge cases, with several
overshooting (per-method Clean-up and Retention per cohort in
Tabs.~\ref{tab:cohort-cleanup},~\ref{tab:cohort-retention},
App.~\ref{app:cohort-cleanup-retention}; underlying \% healed and \%
preserved in the corresponding panels of
Fig.~\ref{fig:recovery-preservation-scatter}).

The Cognition cohort ($n = 25$ damaged) is where the field separates.
Healing remains broadly comparable across non-pathological baselines,
but Retention collapses for every method except DG-Hard and
V-SoftMask: DG-Hard's Cognition Retention of $82.5$ is roughly
$25$\,pp above the next strongest baseline (WiSE-FT at $57.5$) and
substantially higher than L1-reg ($13.0$), while CoFi-Tune, FAPM, and
LoRA collapse to zero because their average \% preserved on the
Cognition partition is non-positive
(Tab.~\ref{tab:cohort-retention}).
\textbf{The Cognition preservation gap, is what separates
DG-Hard from the rest of the field and is the most plausible
mechanism behind its reasoning-model lead.}
V-SoftMask preserves Cognition at near-perfect levels but heals
almost nothing on the same cohort, so it does not function as a
usable repair for any downstream task that requires reasoning.

The same separation appears geometrically in
Fig.~\ref{fig:pareto-cleanup-retention}. Only DG-Hard sits in the
upper-right of both panels; FAPM and V-SoftMask anchor opposite
corners (extreme on one axis at the cost of the other); WiSE-FT
trails DG-Hard but stays closest among the baselines; L1-reg,
CoFi-Tune, and LoRA spread along the lower edges (low Retention on
the left, low Cognition Combined on the right). For repair tasks,
where neither axis can be allowed to collapse, this rules out the
corners and the lower edges alike.

\subsection{Safety alignment recovery}
\label{sec:safety}

Both base models are safety aligned at the Instruct tier, and
benign supervised fine-tuning on knowledge or reasoning data is not
designed to touch alignment. The question is whether the FT updates
we study erode safety as an incidental side effect, and whether
DG-Hard, tuned only on capability metrics, restores it without being
asked to. We score each $\{\text{Base},\text{FT},\text{DG-Hard}\}$
triple on three axes: HarmBench v1.0~\citep{mazeika2024harmbench}
for refusal of harmful prompts, XSTest v2~\citep{rottger2024xstest}
for over-refusal of safe prompts, and StrongREJECT
v1.0~\citep{souly2024strongreject} for continuous-rubric harmfulness
on forbidden prompts (Tab.~\ref{tab:safety-harmbench}).

\begin{table}[t]
  \centering
  \caption{Per-cell Base / FT / DG-Hard scores on three safety axes:
  HarmBench refusal~\citep{mazeika2024harmbench} ($\uparrow$),
  XSTest over-refusal~\citep{rottger2024xstest} ($\downarrow$), and
  StrongREJECT harm~\citep{souly2024strongreject} ($\downarrow$).
  \textbf{Bold} = DG-Hard best among $\{\text{Base},\text{FT},\text{DG}\}$;
  \underline{underline} = second. Eval details in App.~\ref{app:safety-eval}.}
  \label{tab:safety-harmbench}
  \label{tab:safety-xstest}
  \label{tab:safety-strongreject}
  \footnotesize
  \setlength{\tabcolsep}{4pt}
  \begin{tabular}{ll ccc ccc ccc}
  \toprule
  & & \multicolumn{3}{c}{HarmBench refusal $\uparrow$} & \multicolumn{3}{c}{XSTest over-refusal $\downarrow$} & \multicolumn{3}{c}{StrongREJECT harm $\downarrow$} \\
  \cmidrule(lr){3-5}\cmidrule(lr){6-8}\cmidrule(lr){9-11}
  Model & Task & Base & FT & DG & Base & FT & DG & Base & FT & DG \\
  \midrule
  qwen3p5\_4b   & winogrande & 96.0 & 81.0  & \underline{94.5} & 2.8 & 0.8  & \underline{1.2} & 0.0160 & 0.0160 & \textbf{0.0096} \\
  qwen3p5\_4b   & wikiqa     & 97.0 & 100.0 & \underline{97.0} & 2.8 & 15.6 & \underline{4.0} & 0.0160 & 0.0076 & \underline{0.0080} \\
  qwen3p5\_4b   & medqa      & 95.0 & 99.5  & 91.5             & 2.0 & 4.4  & \textbf{1.2}    & 0.0076 & 0.0096 & 0.0128 \\
  llama3p2\_3b  & winogrande & 88.0 & 75.5  & \underline{84.0} & 8.0 & 3.6  & \underline{5.6} & 0.0479 & 0.1402 & \underline{0.0931} \\
  llama3p2\_3b  & wikiqa     & 87.5 & 84.5  & \textbf{87.5}    & 7.6 & 4.4  & \underline{6.8} & 0.0531 & 0.0567 & \textbf{0.0463} \\
  llama3p2\_3b  & medqa      & 87.5 & 79.0  & \underline{84.0} & 7.6 & 4.0  & \underline{5.2} & 0.0539 & 0.0795 & \underline{0.0663} \\
  \bottomrule
  \end{tabular}
\end{table}

The signal-plus-noise model predicts that safety alignment, encoded
in pretrained directions Base already occupies, should survive a
repair that removes only the IID noise residual. The data is
consistent with this prediction. On every Llama cell where FT
lowered HarmBench refusal or raised StrongREJECT harmfulness,
DG-Hard recovers the majority of the gap toward Base. No cell shows
large degradation, including the Qwen cells where FT incidentally
improved a safety axis.

Qwen+wikiqa provides a more demanding test. FT shifts alignment in
opposite directions on the same model: refusal of harmful prompts
saturates at $100\%$, while XSTest over-refusal of safe prompts
rises from $2.8$ to $15.6$. DG-Hard returns both axes to within
small residuals of Base. That a non-safety-specific spectral repair
simultaneously corrects both opposite-direction shifts on the same
cell is consistent with the bidirectional damage sharing the
noise-residual structure DG-Hard targets.

StrongREJECT's continuous rubric provides a finer-grained signal
than the binary refusal label: on Llama, non-refusals are
accompanied by greater per-occurrence severity, and DG-Hard recovers
a substantial fraction of that severity. The threshold was set
without reference to alignment data, so the safety recovery appears
to follow from the noise-residual model the spectral cut is built
around, rather than from a designed objective.

The broader pattern across capability and safety axes is consistent.
DG-Hard occupies a balanced position between Base and FT: damaged
held-out capabilities recover toward Base, on-task gains track FT,
and safety alignment is preserved as an apparent side effect of the
same spectral cut. The three axes appear to track one mechanism
rather than three.


\section{Limitations and conclusion}
\label{sec:limitations}

\paragraph{Limitations.}
DG-Hard treats the task-relevant update as concentrated in a small
number of high-singular-value directions; small-spectral-energy
components are zeroed regardless of semantic importance, which can
fail when the useful signal is spectrally diffuse. This assumption
is consistent with prior work showing that fine-tuning weight
updates are rank-deficient~\citep{hu2022lora}, that the fine-tuning
objective has low intrinsic dimensionality~\citep{aghajanyan2021intrinsic},
and that full-FT updates carry higher effective rank than LoRA
reparameterizations~\citep{shuttleworth2025lora}
(App.~\ref{app:low-rank-evidence}); the failure mode applies to a
minority regime but is not ruled out a priori.

The only operational consideration is the per-matrix SVD cost,
$O(\min(m,n)^{2}\,\max(m,n))$. Even at frontier scale the dominant
matrix grows only $\sim 7\times$ from Llama-3.2-3B's embedding
($128{,}256 \times 3{,}072$) to Kimi-K2.6's ($163{,}840 \times
7{,}168$). At the $\sim 1$T-parameter scale of Kimi-K2.6, the
remaining mixture-of-experts components admit parallel SVD; a full
repair completes in hours, a negligible fraction of the weeks of
training that produced the checkpoints.

\paragraph{Conclusion.}
Fine-tuning weight deltas admit a clean two-component decomposition:
a low-rank task-aligned update and an IID-like noise residual that
gradient descent has no incentive to remove. DG-Hard strips the
residual by applying the Donoho-Gavish optimal hard SVD threshold
post hoc, data-free, and gradient-free. Across $14$ (model, task)
cells on Qwen3.5-4B and Llama-3.2-3B-Instruct and nine cross-domain
held-out benchmarks, DG-Hard achieves the highest balanced point on
the recovery-preservation trade-off under a partition-conditional
Combined metric, beating the next-best post-hoc baseline by
$+6.9$\,pp, and restores safety alignment that benign fine-tuning
incidentally erodes on three independent safety axes without
alignment data entering the procedure. The capability and safety
patterns appear to follow from a single mechanism rather than three.
Additional experiments supporting these design choices, including a
comparison with element-wise model-merging baselines, are in
App.~\ref{app:ablations}.


\clearpage
\bibliographystyle{unsrtnat}
\bibliography{references}

@article{marchenko1967distribution,
  title={Distribution of Eigenvalues for Some Sets of Random Matrices},
  author={Marchenko, V. A. and Pastur, L. A.},
  journal={Mathematics of the USSR-Sbornik},
  volume={1},
  number={4},
  pages={457--483},
  year={1967}
}

@incollection{mccloskey1989catastrophic,
  title={Catastrophic Interference in Connectionist Networks: The Sequential Learning Problem},
  author={McCloskey, Michael and Cohen, N. J.},
  booktitle={Psychology of Learning and Motivation},
  volume={24},
  pages={109--165},
  year={1989},
  publisher={Academic Press}
}

@inproceedings{keskar2016large,
  title={On Large-Batch Training for Deep Learning: Generalization Gap and Sharp Minima},
  author={Keskar, N. S. and Mudigere, Dheevatsa and Nocedal, Jorge and Smelyanskiy, Misha and Tang, Ping Tak Peter},
  booktitle={International Conference on Learning Representations},
  year={2017},
  eprint={1609.04836},
  archivePrefix={arXiv},
  primaryClass={stat.ML}
}

@article{jastrzebski2017three,
  title={Three Factors Influencing Minima in {SGD}},
  author={Jastrz{\k{e}}bski, Stanis{\l}aw and Kenton, Zachary and Arpit, Devansh and Ballas, Nicolas and Fischer, Asja and Bengio, Yoshua and Storkey, Amos},
  journal={arXiv preprint arXiv:1711.04623},
  year={2017}
}

@inproceedings{kumar2022finetuning,
  title={Fine-Tuning Can Distort Pretrained Features and Underperform Out-of-Distribution},
  author={Kumar, Ananya and Raghunathan, Aditi and Jones, Robbie and Ma, Tengyu and Liang, Percy},
  booktitle={International Conference on Learning Representations},
  year={2022},
  eprint={2202.10054},
  archivePrefix={arXiv},
  primaryClass={cs.LG}
}

@inproceedings{aghajanyan2021intrinsic,
  title={Intrinsic Dimensionality Explains the Effectiveness of Language Model Fine-Tuning},
  author={Aghajanyan, Armen and Zettlemoyer, Luke and Gupta, Sonal},
  booktitle={Proceedings of the 59th Annual Meeting of the Association for Computational Linguistics},
  year={2021},
  url={https://arxiv.org/abs/2012.13255}
}

@inproceedings{aljundi2018memory,
  title={Memory Aware Synapses: Learning What (not) to Forget},
  author={Aljundi, Rahaf and Babiloni, Francesca and Elhoseiny, Mohamed and Rohrbach, Marcus and Tuytelaars, Tinne},
  booktitle={Proceedings of the European Conference on Computer Vision},
  year={2018},
  url={https://arxiv.org/abs/1711.09601}
}

@inproceedings{brown2020language,
  title={Language Models are Few-Shot Learners},
  author={Brown, Tom B. and Mann, Benjamin and Ryder, Nick and Subbiah, Melanie and Kaplan, Jared and Dhariwal, Prafulla and Neelakantan, Arvind and Shyam, Pranav and Sastry, Girish and Askell, Amanda and Agarwal, Sandhini and Herbert-Voss, Ariel and Krueger, Gretchen and Henighan, Tom and Child, Rewon and Ramesh, Aditya and Ziegler, Daniel M. and Wu, Jeffrey and Winter, Clemens and Hesse, Christopher and Chen, Mark and Sigler, Eric and Litwin, Mateusz and Gray, Scott and Chess, Benjamin and Clark, Jack and Berner, Christopher and McCandlish, Sam and Radford, Alec and Sutskever, Ilya and Amodei, Dario},
  booktitle={Advances in Neural Information Processing Systems},
  year={2020},
  url={https://arxiv.org/abs/2005.14165}
}

@article{chaudhry2019tiny,
  title={On Tiny Episodic Memories in Continual Learning},
  author={Chaudhry, Arslan and Rohrbach, Marcus and Elhoseiny, Mohamed and Ajanthan, Thalaiyasingam and Dokania, Puneet K. and Torr, Philip H. S. and Ranzato, Marc'Aurelio},
  journal={arXiv preprint arXiv:1902.10486},
  year={2019},
  url={https://arxiv.org/abs/1902.10486}
}

@inproceedings{devlin2019bert,
  title={{BERT}: Pre-training of Deep Bidirectional Transformers for Language Understanding},
  author={Devlin, Jacob and Chang, Ming-Wei and Lee, Kenton and Toutanova, Kristina},
  booktitle={Proceedings of the 2019 Conference of the North American Chapter of the Association for Computational Linguistics: Human Language Technologies},
  year={2019},
  url={https://arxiv.org/abs/1810.04805}
}

@article{donoho2014optimal,
  title={The Optimal Hard Threshold for Singular Values is $4/\sqrt{3}$},
  author={Gavish, Matan and Donoho, David L.},
  journal={IEEE Transactions on Information Theory},
  volume={60},
  number={8},
  pages={5040--5053},
  year={2014},
  url={https://arxiv.org/abs/1305.5870}
}

@article{french1999catastrophic,
  title={Catastrophic Forgetting in Connectionist Networks},
  author={French, Robert M.},
  journal={Trends in Cognitive Sciences},
  volume={3},
  number={4},
  pages={128--135},
  year={1999},
  doi={10.1016/S1364-6613(99)01294-2}
}

@article{gavish2017optimal,
  title={Optimal Shrinkage of Singular Values},
  author={Gavish, Matan and Donoho, David L.},
  journal={IEEE Transactions on Information Theory},
  volume={63},
  number={4},
  pages={2137--2152},
  year={2017},
  url={https://arxiv.org/abs/1405.7511}
}

@inproceedings{howard2018universal,
  title={Universal Language Model Fine-tuning for Text Classification},
  author={Howard, Jeremy and Ruder, Sebastian},
  booktitle={Proceedings of the 56th Annual Meeting of the Association for Computational Linguistics},
  year={2018},
  url={https://arxiv.org/abs/1801.06146}
}

@inproceedings{hu2022lora,
  title={{LoRA}: Low-Rank Adaptation of Large Language Models},
  author={Hu, Edward J. and Shen, Yelong and Wallis, Phillip and Allen-Zhu, Zeyuan and Li, Yuanzhi and Wang, Shean and Wang, Lu and Chen, Weizhu},
  booktitle={International Conference on Learning Representations},
  year={2022},
  url={https://arxiv.org/abs/2106.09685}
}

@inproceedings{huang2025fapm,
  title={Mitigating Catastrophic Forgetting in Large Language Models with Forgetting-aware Pruning ({FAPM})},
  author={Huang, Wei and Cheng, Aimin and Wang, Yu},
  booktitle={Proceedings of the 2025 Conference on Empirical Methods in Natural Language Processing},
  year={2025},
  url={https://arxiv.org/abs/2509.08255}
}

@inproceedings{ilharco2023editing,
  title={Editing Models with Task Arithmetic},
  author={Ilharco, Gabriel and Ribeiro, Marco Tulio and Wortsman, Mitchell and Gururangan, Suchin and Schmidt, Ludwig and Hajishirzi, Hannaneh and Farhadi, Ali},
  booktitle={International Conference on Learning Representations},
  year={2023},
  url={https://arxiv.org/abs/2212.04089}
}

@inproceedings{ke2023continual,
  title={Continual Pre-Training of Language Models},
  author={Ke, Zixuan and Shao, Yijia and Lin, Haolong and Konishi, Tatsuya and Kim, Gyuwan and Liu, Bing},
  booktitle={International Conference on Learning Representations},
  year={2023},
  url={https://arxiv.org/abs/2302.03241}
}

@article{kirkpatrick2017overcoming,
  title={Overcoming Catastrophic Forgetting in Neural Networks},
  author={Kirkpatrick, James and Pascanu, Razvan and Rabinowitz, Neil and Veness, Joel and Desjardins, Guillaume and Rusu, Andrei A. and Milan, Kieran and Quan, John and Ramalho, Tiago and Grabska-Barwi{\'n}ska, Agnieszka and Hassabis, Demis and Clopath, Claudia and Kumaran, Dharshan and Hadsell, Raia},
  journal={Proceedings of the National Academy of Sciences},
  volume={114},
  number={13},
  pages={3521--3526},
  year={2017},
  url={https://arxiv.org/abs/1612.00796}
}

@inproceedings{lopez2017gradient,
  title={Gradient Episodic Memory for Continual Learning},
  author={Lopez-Paz, David and Ranzato, Marc'Aurelio},
  booktitle={Advances in Neural Information Processing Systems},
  year={2017},
  url={https://arxiv.org/abs/1706.08840}
}

@article{luo2023empirical,
  title={An Empirical Study of Catastrophic Forgetting in Large Language Models During Continual Fine-tuning},
  author={Luo, Yun and Yang, Zefan and Meng, Fan and Li, Yufei and Zhou, Jie and Zhang, Yue},
  journal={arXiv preprint arXiv:2308.08747},
  year={2023},
  url={https://arxiv.org/abs/2308.08747}
}

@inproceedings{matena2022merging,
  title={Merging Models with Fisher-Weighted Averaging},
  author={Matena, Michael S. and Raffel, Colin A.},
  booktitle={Advances in Neural Information Processing Systems},
  year={2022},
  url={https://arxiv.org/abs/2111.09832}
}

@inproceedings{mazeika2024harmbench,
  title={{HarmBench}: A Standardized Evaluation Framework for Automated Red Teaming and Robust Refusal},
  author={Mazeika, Mantas and Phan, Long and Yin, Xuwang and Zou, Andy and Wang, Zifan and Mu, Norman and Sakhaee, Elham and Li, Nathaniel and Basart, Steven and Li, Bo and Forsyth, David and Hendrycks, Dan},
  booktitle={Proceedings of the 41st International Conference on Machine Learning},
  year={2024},
  url={https://arxiv.org/abs/2402.04249}
}

@inproceedings{ouyang2022training,
  title={Training Language Models to Follow Instructions with Human Feedback},
  author={Ouyang, Long and Wu, Jeffrey and Jiang, Xu and Almeida, Diogo and Wainwright, Carroll and Mishkin, Pamela and Zhang, Chong and Agarwal, Sandhini and Slama, Katarina and Ray, Alex and Schulman, John and Hilton, Jacob and Kelton, Fraser and Miller, Luke and Simens, Maddie and Askell, Amanda and Welinder, Peter and Christiano, Paul and Leike, Jan and Lowe, Ryan},
  booktitle={Advances in Neural Information Processing Systems},
  year={2022},
  url={https://arxiv.org/abs/2203.02155}
}

@inproceedings{panigrahi2023task,
  title={Task-Specific Skill Localization in Fine-Tuned Language Models},
  author={Panigrahi, Abhishek and Saunshi, Nikunj and Zhao, Haifeng and Arora, Sanjeev},
  booktitle={International Conference on Machine Learning},
  year={2023},
  url={https://arxiv.org/abs/2302.06600}
}

@inproceedings{qi2024finetuning,
  title={Fine-Tuning Aligned Language Models Compromises Safety, Even When Users Do Not Intend To!},
  author={Qi, Xiangyu and Zeng, Yi and Xie, Tinghao and Chen, Pin-Yu and Jia, Ruoxi and Mittal, Prateek and Henderson, Peter},
  booktitle={International Conference on Learning Representations},
  year={2024},
  url={https://arxiv.org/abs/2310.03693}
}

@article{ratcliff1990connectionist,
  title={Connectionist Models of Recognition Memory: Constraints Imposed by Learning and Forgetting Functions},
  author={Ratcliff, Roger},
  journal={Psychological Review},
  volume={97},
  number={2},
  pages={285-308},
  year={1990},
  doi={10.1037/0033-295X.97.2.285}
}

@inproceedings{rottger2024xstest,
  title={{XSTest}: A Test Suite for Identifying Exaggerated Safety Behaviours in Large Language Models},
  author={R{\"o}ttger, Paul and Kirk, Hannah Rose and Vidgen, Bertie and Attanasio, Giuseppe and Bianchi, Federico and Hovy, Dirk},
  booktitle={Proceedings of the 2024 Conference of the North American Chapter of the Association for Computational Linguistics},
  year={2024},
  url={https://arxiv.org/abs/2308.01263}
}

@inproceedings{sharma2023truth,
  title={The Truth is in There: Improving Reasoning in Language Models with Layer-Selective Rank Reduction},
  author={Sharma, Pratyusha and Ash, Jordan T. and Misra, Dipendra},
  booktitle={International Conference on Learning Representations},
  year={2024},
  url={https://arxiv.org/abs/2312.13558}
}

@inproceedings{shuttleworth2025lora,
  title={{LoRA} vs Full Fine-tuning: An Illusion of Equivalence},
  author={Shuttleworth, Richard and Andreas, Jacob and Torralba, Antonio and Sharma, Pratyusha},
  booktitle={Advances in Neural Information Processing Systems},
  year={2025},
  url={https://arxiv.org/abs/2410.21228}
}

@inproceedings{souly2024strongreject,
  title={A {StrongREJECT} for Empty Jailbreaks},
  author={Souly, Alexandra and Lu, Qingyuan and Bowen, Dillon and Trinh, Tu and Hsieh, Elvis and Pandey, Sana and Abbeel, Pieter and Svegliato, Justin and Emmons, Scott and Watkins, Olivia and Toyer, Sam},
  booktitle={Advances in Neural Information Processing Systems (Datasets and Benchmarks Track)},
  year={2024},
  url={https://arxiv.org/abs/2402.10260}
}

@article{staats2023boundary,
  title={Boundary between noise and information applied to filtering neural network weight matrices},
  author={Staats, Max and Thamm, Matthias and Rosenow, Bernd},
  journal={Physical Review E},
  volume={108},
  pages={L022302},
  year={2023},
  url={https://arxiv.org/abs/2206.03927}
}

@article{thamm2022random,
  title={Random matrix analysis of deep neural network weight matrices},
  author={Thamm, Matthias and Staats, Max and Rosenow, Bernd},
  journal={Physical Review E},
  volume={106},
  pages={054124},
  year={2022},
  url={https://arxiv.org/abs/2203.14661}
}

@inproceedings{wortsman2022robust,
  title={Robust Fine-Tuning of Zero-Shot Models},
  author={Wortsman, Mitchell and Ilharco, Gabriel and Kim, Jong Wook and Li, Mike Y. and Kornblith, Simon and Roelofs, Rebecca and Lopes, Raphael Gontijo and Hajishirzi, Hannaneh and Farhadi, Ali and Namkoong, Hongseok and Schmidt, Ludwig},
  booktitle={Proceedings of the IEEE/CVF Conference on Computer Vision and Pattern Recognition},
  year={2022},
  url={https://arxiv.org/abs/2109.01903}
}

@inproceedings{yadav2023ties,
  title={{TIES}-Merging: Resolving Interference When Merging Models},
  author={Yadav, Prateek and Tam, Derek and Choshen, Leshem and Raffel, Colin and Bansal, Mohit},
  booktitle={Advances in Neural Information Processing Systems},
  year={2023},
  url={https://arxiv.org/abs/2306.01708}
}

@inproceedings{yu2023language,
  title={Language Models are Super Mario: Absorbing Abilities from Homologous Models as a Free Lunch},
  author={Yu, Le and Yu, Bowen and Yu, Haiyang and Huang, Fei and Li, Yong},
  booktitle={International Conference on Machine Learning},
  year={2024},
  url={https://arxiv.org/abs/2311.03099}
}

@inproceedings{zenke2017continual,
  title={Continual Learning Through Synaptic Intelligence},
  author={Zenke, Friedemann and Poole, Ben and Ganguli, Surya},
  booktitle={International Conference on Machine Learning},
  year={2017},
  url={https://arxiv.org/abs/1703.04200}
}

@inproceedings{zhang2024cofitune,
  title={Balancing Speciality and Versatility: A Coarse to Fine Framework for Mitigating Catastrophic Forgetting in Large Language Models},
  author={Zhang, Haonan and Wu, Yinjun and Li, Dongxu and Yang, Shuo and Zhao, Rui and Jiang, Yu and Tan, Fei},
  booktitle={Proceedings of the 62nd Annual Meeting of the Association for Computational Linguistics},
  year={2024},
  url={https://arxiv.org/abs/2404.10306}
}

\newpage
\appendix
\section*{Appendix}

\noindent\textbf{Contents}
\vspace{6pt}

\noindent A\quad \hyperref[app:low-rank-evidence]{Empirical evidence for the signal-plus-noise structure of $\Delta$} \dotfill \pageref{app:low-rank-evidence} \\
B\quad \hyperref[app:noise-scale]{Noise-scale derivation} \dotfill \pageref{app:noise-scale} \\
C\quad \hyperref[app:metric-discussion]{Metric discussion} \dotfill \pageref{app:metric-discussion} \\
D\quad \hyperref[app:setup]{Detailed experimental setup} \dotfill \pageref{app:setup} \\
\hspace*{1.4em}D.1\quad \hyperref[app:setup-tasks]{Fine-tuning tasks} \dotfill \pageref{app:setup-tasks} \\
\hspace*{1.4em}D.2\quad \hyperref[app:setup-benchmarks]{Held-out benchmarks} \dotfill \pageref{app:setup-benchmarks} \\
\hspace*{1.4em}D.3\quad \hyperref[app:setup-hparams]{Training hyperparameters} \dotfill \pageref{app:setup-hparams} \\
\hspace*{1.4em}D.4\quad \hyperref[app:repair-scope]{Repair scope} \dotfill \pageref{app:repair-scope} \\
E\quad \hyperref[app:per-cell-tables]{Per-cell forgetting and balance scores} \dotfill \pageref{app:per-cell-tables} \\
F\quad \hyperref[app:cohort-cleanup-retention]{Per-cohort Clean-up and Retention} \dotfill \pageref{app:cohort-cleanup-retention} \\
G\quad \hyperref[app:cohort-full]{Full per-cohort sub-score breakdown} \dotfill \pageref{app:cohort-full} \\
H\quad \hyperref[app:percell-bench]{Per-(model, task, method, benchmark) results} \dotfill \pageref{app:percell-bench} \\
\hspace*{1.4em}H.1\quad \hyperref[app:method-configs]{Method-specific configurations} \dotfill \pageref{app:method-configs} \\
\hspace*{1.4em}H.2\quad \hyperref[app:inference]{Inference engine and decoding} \dotfill \pageref{app:inference} \\
I\quad \hyperref[app:ablations]{Other experiments and ablations} \dotfill \pageref{app:ablations} \\
\hspace*{1.4em}I.1\quad \hyperref[app:abl-noise-heatmap]{Layer-wise noise concentration} \dotfill \pageref{app:abl-noise-heatmap} \\
\hspace*{1.4em}I.2\quad \hyperref[app:abl-mask]{Layer-mask causal test} \dotfill \pageref{app:abl-mask} \\
\hspace*{1.4em}I.3\quad \hyperref[app:abl-merging]{Comparison with element-wise merging baselines} \dotfill \pageref{app:abl-merging}

\newpage

\section{Empirical evidence for the signal-plus-noise structure of $\Delta$}
\label{app:low-rank-evidence}

\begin{figure}[t]
\centering
\includegraphics[width=\textwidth]{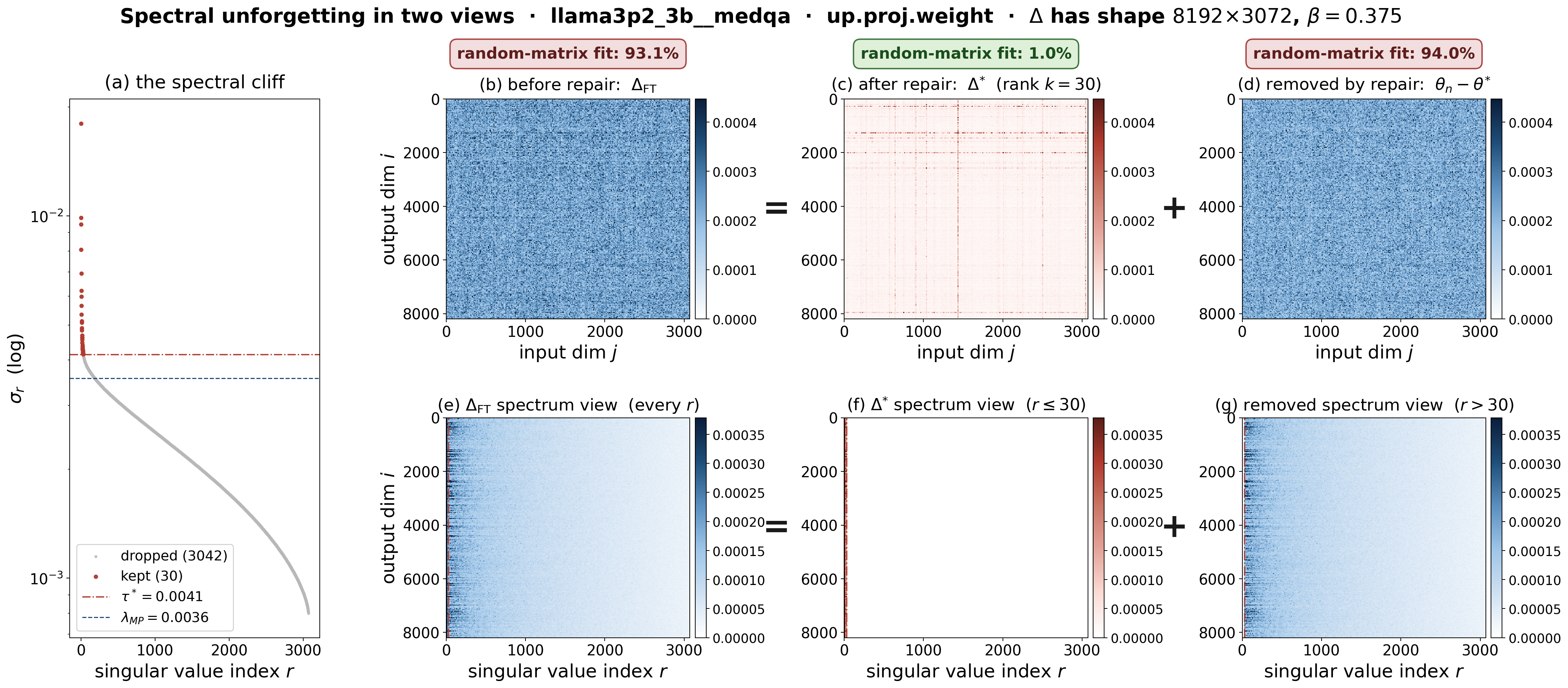}
\caption{Spectral unforgetting in two views, on Llama-3.2-3B
\texttt{mlp.up\_proj} at layer~$14$, with $\Delta = \Wft - \Wbase \in
\mathbb{R}^{8192 \times 3072}$ ($\beta = 0.375$). \textbf{(a)} The
fine-tune delta has a sharp spectral cliff: $30$ singular values
(red) lie above the DG-Hard threshold $\tau^{\ast} =
\omega(\beta)\hat{\sigma}$, while the remaining $3042$ (gray) sit at
or below the Marchenko-Pastur bulk edge $\lammp$. \textbf{Top row,
(b) to (d): entry-space view} of the additive identity
$\Delta_{\mathrm{FT}} = \Delta^{\ast} + (W_{\mathrm{ft}} - W^{\ast})$,
with each pixel showing $\max|\Delta_{ij}|$ over a block of matrix
entries. The full FT delta (b) is uniform speckle; the rank-$30$
repaired delta (c) reveals horizontal banding from the kept
left-singular vectors $u_r$; the discarded component (d) is again
uniform speckle, carrying no spatial structure. \textbf{Bottom row,
(e) to (g): spectrum-space view} of the same identity, with each
pixel showing $\max \sigma_r |u_r[i]|$, the contribution of singular
direction $r$ to output neuron $i$. Panel (e) decomposes into the
kept-only panel (f), nonzero only in its leftmost $30$ columns, and
the bulk-only panel (g), nonzero everywhere except those columns;
their pixel-wise sum reproduces (e) exactly, because every singular
direction belongs to exactly one set. Random-matrix-fit chips above
the top row report the percentage match between each matrix's
singular-value distribution and the Marchenko-Pastur prediction
(green: structured, non-random; red: IID-noise-like):
$\Delta_{\mathrm{FT}}$ scores $93.1\%$, the rank-$30$ repaired delta
$\Delta^{\ast}$ scores $1.0\%$, and the discarded component scores
$94.0\%$, slightly more MP-like than $\Delta_{\mathrm{FT}}$ itself,
confirming that DG-Hard separates the two without leaving residual
signal in the noise.}
\label{fig:spectral-unforgetting}
\end{figure}

The central modeling assumption of selective spectral reversion is
that, for each $2$D weight matrix $W$ in the network, the
fine-tuning delta $\Delta = \Wft - \Wbase$ (equivalently $\Delta W$
in much of the prior LoRA-adjacent literature) admits a
two-component decomposition $\Delta = \Delta_{\mathrm{signal}} +
\Delta_{\mathrm{noise}}$, where $\Delta_{\mathrm{signal}}$ is a
low-rank task-aligned update and $\Delta_{\mathrm{noise}}$ is a
residual that behaves statistically like an IID random matrix. The
phenomenological evidence the model is designed to explain is the
per-cell forgetting pattern in Tab.~\ref{tab:cell-damage}: every
fine-tune produces a small set of large per-benchmark drops on a
small set of held-out benchmarks (concentrated signal-side damage),
while the remaining held-out scores drift only mildly (diffuse
noise-side damage). Two independent threads of empirical evidence,
one from the fine-tuning literature and one from the
random-matrix-theory literature, support this picture, and a direct
check on our own checkpoints (Fig.~\ref{fig:spectral-unforgetting})
completes the argument.

\paragraph{The signal side: the task-aligned update is
low-dimensional.}
\citet{aghajanyan2021intrinsic} establish low intrinsic
dimensionality for the fine-tuning objective by showing that
optimizing only $\sim 200$ trainable parameters via random projection
into the full parameter space recovers $90\%$ of full-fine-tune
performance on MRPC with RoBERTa; this is a property of the
optimization landscape rather than of the realized weight delta.
\citet{hu2022lora} provide the more direct evidence: in their
intrinsic-rank analysis, very low LoRA ranks (often single-digit)
match high-rank LoRA on downstream tasks, and the learned $\Delta W$
amplifies a small set of task-specific directions, supporting their
explicit ``rank-deficiency of $\Delta W$'' conclusion.
\citet{shuttleworth2025lora} report that full fine-tuning carries a
higher effective rank than LoRA reparameterizations even at matched
nominal rank, with full-FT deltas modifying the existing pretrained
singular structure and LoRA introducing a small number of additional
``intruder'' directions approximately orthogonal to that structure;
the spike-versus-bulk reading of our signal-plus-noise model survives
both regimes because it isolates the structured component above the
MP bulk rather than committing to a strict low-rank cap.
\citet{panigrahi2023task} provide a complementary parametric-sparsity
result: $\sim 0.01\%$ of parameters carry $> 95\%$ of fine-tune task
performance when grafted back onto the base model. Sparsity in
coordinate space and concentration in singular-value space are
distinct mathematical properties; we cite this work as a parallel
structural prior, not as direct evidence for low-rank $\Delta$.

\paragraph{The noise side: the residual past the spikes follows the
Marchenko-Pastur bulk.} Random-matrix theory is an empirically
validated description of trained neural-network weight matrices, not
just an asymptotic abstraction. \citet{thamm2022random} analyze the
singular spectra of trained deep-network weight matrices and
demonstrate that the bulk fits the MP density layer by layer, with a
finite number of outliers carrying the learned signal.
\citet{staats2023boundary} operationalize the same picture: they
apply MP-edge filtering directly to neural-network weight matrices,
set the sub-edge bulk to zero, and recover the trained network's
behavior, providing direct evidence that the bulk is functionally
inert. Two independent fine-tuning-side observations point in the
same direction. \citet{sharma2023truth} (LASER) show that
aggressively zeroing the high-rank tail of a trained transformer's
weight matrix can improve downstream performance, suggesting the
high-rank bulk encodes noise rather than usable signal, a result
that transfers to the fine-tune delta by linearity of SVD truncation.
\citet{yu2023language} (DARE) show that the fine-tune delta tolerates
random pruning of $90$--$99\%$ of its entries with rescaling,
attributing this to ``extreme redundancy'' of small-magnitude
updates, consistent with most of $\Delta$ being redundant rather than
informative.

\paragraph{Direct check on our own checkpoints.}
Fig.~\ref{fig:spectral-unforgetting} verifies the structure
layer-locally on Llama-3.2-3B's \texttt{mlp.up\_proj} at layer~$14$.
Panel~(a) shows the spectral cliff at the DG-Hard threshold
$\tau^{\ast}$. Panels (b) to (g) show that the additive identity
$\Delta_{\mathrm{FT}} = \Delta^{\ast} + (W_{\mathrm{ft}} - W^{\ast})$
holds simultaneously in entry space (top row) and in
singular-vector space (bottom row). The random-matrix-fit chips on
the top row verify the decomposition quantitatively. We define
\begin{equation}
\label{eq:mp-fit}
\mathrm{MP\text{-}fit}(\mathbf{s})
\;=\;
\bigl(1 - \mathrm{KS}(F_{\mathbf{s}},\, F_{\mathrm{MP}}(\hat\sigma, \beta))\bigr)
\cdot 100,
\end{equation}
where $F_{\mathbf{s}}$ is the empirical CDF of the matrix's singular
values, $F_{\mathrm{MP}}$ is the closed-form Marchenko-Pastur CDF at
the noise-scale estimate $\hat\sigma$ shared across the three
matrices and the aspect ratio $\beta$, and $\mathrm{KS}$ is the
Kolmogorov-Smirnov distance between them; $\mathrm{MP\text{-}fit} = 100$
indicates that the empirical and theoretical CDFs coincide
($\mathbf{s}$ is statistically indistinguishable from IID noise),
while $\mathrm{MP\text{-}fit} \to 0$ indicates a maximally non-random
spectrum. Under this metric, $\Delta_{\mathrm{FT}}$ scores $93.1\%$,
the rank-$30$ repaired delta $\Delta^{\ast}$ scores $1.0\%$
(explicitly non-random), and the discarded component scores
$94.0\%$, slightly more MP-like than $\Delta_{\mathrm{FT}}$ itself,
confirming that DG-Hard extracts the structured component without
leaving residual signal in the noise.

\paragraph{Where this lands the central assumption.}
The signal-side and noise-side evidence jointly support the
decomposition: the weight update is rank-deficient
\citep{hu2022lora} and the fine-tuning objective has low intrinsic
dimensionality \citep{aghajanyan2021intrinsic}, with related findings
on the spectral structure of full-FT versus LoRA deltas
\citep{shuttleworth2025lora} and on parametric sparsity
\citep{panigrahi2023task}; the residual is empirically
MP-bulk-distributed \citep{thamm2022random, staats2023boundary}, and
that bulk is functionally redundant under direct manipulation of
trained weights and fine-tune deltas \citep{sharma2023truth,
yu2023language}.
Fig.~\ref{fig:spectral-unforgetting} checks the bulk-fit and
additivity claims on the exact $\Delta$ matrix the method operates
on.

\FloatBarrier

\section{Noise-scale derivation}
\label{app:noise-scale}

\S\ref{sec:dghard} of the main paper introduces the Donoho-Gavish
noise estimator. This appendix gives the full
derivation and the implementation footprint.

Equation~\eqref{eq:dg-hard-tau} requires the noise scale $\sigma$ of
the residual $\Delta_{\mathrm{noise}}$ in \eqref{eq:s+n}. Ideally we
would estimate $\sigma$ from $\Delta_{\mathrm{noise}}$ directly, but
the decomposition in \eqref{eq:s+n} is unobservable.
\citet{donoho2014optimal} resolve this within the same MP framework.
Under the spike model, only the top $r$ singular values carry signal
contamination, with $r$ unknown but small relative to $p$; the
remaining $p - r$ are asymptotically distributed as the
singular-value MP density at aspect $\beta$. That density has a
closed-form median $\mu_{\beta}$,
\begin{equation}
\label{eq:mu-beta}
\mu_{\beta} \;=\; \sqrt{\med(f_{\beta})},
\qquad
f_{\beta}(\lambda)
\;=\; \frac{\sqrt{(\lambda_{+} - \lambda)(\lambda - \lambda_{-})}}
            {2\pi\,\beta\,\lambda},
\quad
\lambda_{\pm} = (1 \pm \sqrt{\beta})^{2},
\end{equation}
which we precompute by fine-grid trapezoidal integration once per
distinct aspect ratio in the network. Because $r \ll p$, the median
of the \emph{empirical} singular values $\med(\mathbf{s})$ falls
inside the noise bulk regardless of the precise value of $r$, and
asymptotically tracks $\sigma\sqrt{\max(m, n)}\cdot\mu_{\beta}$.
Solving for $\sigma$ gives the Donoho-Gavish noise estimator,
\begin{equation}
\label{eq:sighat}
\sighat \;=\; \frac{\med(\mathbf{s})}{\mu_{\beta}\,\sqrt{\max(m,n)}}.
\end{equation}
Equation~\eqref{eq:sighat} is internally consistent with the rest of
the DG framework: the same MP density that determines the bulk edge
\eqref{eq:mp-edge} and the optimal threshold \eqref{eq:dg-hard-tau}
also determines $\mu_{\beta}$. Substituting \eqref{eq:sighat} into
\eqref{eq:dg-hard-tau} eliminates $\sqrt{\max(m,n)}$ and yields the
unconditional form of the DG hard threshold,
\begin{equation}
\label{eq:dg-hard-data}
\tau^{*} \;=\; \frac{\omega(\beta)}{\mu_{\beta}}\,\med(\mathbf{s}),
\end{equation}
which evaluates to $\approx 2.858 \cdot \med(\mathbf{s})$ for square
matrices and is computable directly from the empirical spectrum
without any auxiliary statistic.

\paragraph{Cost and implementation footprint.}
The per-matrix dominant cost is the SVD on line~\ref{algline:sighat}'s
preceding step, $O(\min(m,n)^{2}\,\max(m,n))$, completing in a few
hundred milliseconds for the largest matrix in the model sizes we
evaluate (the Llama-3.2-3B-Instruct embedding,
$128{,}256 \times 3{,}072$); full-network repair completes in
$\le 3$ minutes on a single A$100$-$80$GB for the model sizes we evaluate. The
implementation loads $\Wbase$ and $\Wft$ once, processes
Algorithm~\ref{alg:dg-repair} tensor-by-tensor, and stitches repaired
matrices back at their original keys. The repair scope is restricted
to parameters with $\mathrm{ndim} \ge 2$ and at least $1024$ elements;
tensors with $\mathrm{ndim} > 2$ (e.g., $1$D convolution kernels) are
reshaped to $2$D before SVD and reshaped back afterwards. $1$D
parameters (biases, normalization scales) are left at their
fine-tuned values since they admit no non-trivial MP edge. Peak GPU memory is approximately twice the model's parameter
footprint plus a working buffer for the batched SVD, well within a
$40$\,GB A$100$.

\section{Metric discussion}
\label{app:metric-discussion}

\paragraph{\% healed.}
Conventional held-out accuracy averages across all benchmarks
regardless of whether fine-tuning damaged them, so a method that
simply mirrors FT scores indistinguishably from one that actively
repairs. \% healed restricts the average to the FT-damaged partition
$D$ and reports how much of the FT-to-Base gap each method closes.
The numerator is the repaired score's gain over FT; the denominator
is the gap fine-tuning opened, so $0$ means no recovery and $100$
means full restoration to Base.

\paragraph{\% preserved.}
The analogous construction on the FT-improved partition $I$. A flat
held-out average pools FT-improved triples with FT-unchanged ones,
so a method that erases every incidental gain looks numerically
similar to one that keeps them whenever most cells were FT-untouched.
\% preserved isolates the triples where fine-tuning produced a real
lift and reports the share of that lift the repaired model still
carries.

\paragraph{Non-damage.}
Repairs that aggressively shrink the delta can introduce regressions
on benchmarks fine-tuning never touched. A flat held-out mean
smooths these regressions into the average; non-damage surfaces them.
The statistic is the fraction of FT-unchanged triples on which the
repaired score is within $3$\,pp of Base, so a method that bleeds
outside the FT-damaged cells loses points here independently
of how well it heals.

\paragraph{On-task retention.}
Absolute on-task accuracy is misleading: tasks vary in difficulty,
and a flat report rewards methods that simply leave the FT model
unchanged. On-task ret.\ normalizes the method's target-task score
by FT's, so $100$ means every fine-tuning gain on the target task is
preserved and $0$ means none is.

\paragraph{Clean-up and Retention.}
Clean-up $= \mathrm{HM}(\text{\% healed},\,\text{non-damage})$
summarizes the cleaning side: actively reverting FT-damage on the
FT-damaged set $D$ and leaving the FT-unchanged set $U$ alone.
Retention $= \mathrm{HM}(\text{\% preserved},\,\text{on-task ret.})$
summarizes the preservation side: keeping incidental held-out gains
on the FT-improved set $I$ and the target-task gain. Each is a
harmonic mean by design, so a method that maximizes one sub-statistic
by collapsing the other is bottlenecked rather than rewarded;
Combined applies the same bottleneck across the two sides.

\paragraph{Why not a flat unified average?}
A common alternative aggregation in the continual-learning literature,
the Uni.\ score of \citet{zhang2024cofitune}, computes a flat average
of held-out scores and combines it with on-task accuracy. Its
held-out term is blind to what fine-tuning did to each benchmark, so
a method that actively pulls damaged benchmarks back toward base and
a method that simply avoids perturbing the FT distribution can post
numerically similar Uni.\ scores even though only the first is doing
the work the metric implicitly claims to measure. Our scoring
conditions every held-out measurement on its FT classification
(\eqref{eq:delta-defs}): \% healed is averaged only over damaged
cases, \% preserved only over improved ones, so
\emph{recovered-the-damage}
and \emph{never-touched-it} become two distinct statistics rather
than two paths to the same number. Combined with the harmonic-mean
bottleneck of \S\ref{sec:setup-eval}, this gives the metric a
guarantee an unconditional flat sum cannot match.

\paragraph{Per-cell aggregation.}
The Combined score in \eqref{eq:combined-def} is for population-level
(cohort) comparisons. Per-cell scores in Tab.~\ref{tab:cell-balance}
of the main paper use a separate aggregation: the harmonic mean of
$(\bar{s}_{\mathrm{method}}^{\mathrm{held}} /
\bar{s}_{\mathrm{base}}^{\mathrm{held}}) \times 100$ and
$(s_{\mathrm{method}}^{\mathrm{task}} /
s_{\mathrm{FT}}^{\mathrm{task}}) \times 100$, where
$\bar{s}^{\mathrm{held}}$ denotes the mean held-out score across the
cell's nine held-out benchmarks. Each ratio multiplies
by $100$ so that values cleanly exceed $100$ when a method outperforms
its reference (e.g., DG-Hard scores $108.5$ on Qwen3.5-4B + BoolQ).
Per-cell mean ratios are used here in place of the continuous \%
healed / \% preserved statistics because, on $n \approx 9$ benchmarks
per cell, the binary damaged / improved / unchanged classification
produces small-sample sub-statistics that are too noisy to read
cell-by-cell.

\section{Detailed experimental setup}
\label{app:setup}

This appendix documents the reproducibility-level configuration that
\S\ref{sec:experiments} of the main paper summarizes in prose.

\subsection{Fine-tuning tasks}
\label{app:setup-tasks}

Tab.~\ref{tab:tasks} lists the seven fine-tuning tasks forming the row
axis of the experimental matrix, with full HuggingFace dataset
identifiers and eval split sizes. RTE, StrategyQA, ReClor, MedQA, and
Winogrande use their full eval splits; BoolQ and WikiQA are subsetted
to $1500$ items (out of $3270$ and $6165$ respectively) to keep the
$14 \times 9$ matrix wall-clock tractable, with per-method
standard-error well below $1$\,pp at this sample size.

\begin{table}[h]
\centering
\caption{The seven fine-tuning tasks forming the row axis of the
experimental matrix. The eval split of each task becomes that cell's
on-task benchmark \texttt{task\_\{name\}} throughout the analysis.}
\label{tab:tasks}
\small
\setlength{\tabcolsep}{4pt}
\resizebox{\textwidth}{!}{%
\begin{tabular}{@{}llllrl@{}}
\toprule
Task & HF dataset & Train split & Eval split & Eval $n$ &
Reasoning style \\
\midrule
RTE         & \texttt{aps/super\_glue (rte)}   & train          & validation & 277  & binary entailment (2-class) \\
StrategyQA  & \texttt{ChilleD/StrategyQA}      & train          & test       & 687  & implicit multi-hop yes/no \\
ReClor      & \texttt{metaeval/reclor}         & train ($\sim 4.6$k) & validation & 500  & logical 4-way MC \\
BoolQ       & \texttt{aps/super\_glue (boolq)} & train (sub.)   & validation & 1500 & reading comp.\ yes/no \\
MedQA       & \texttt{GBaker/MedQA-USMLE-4-options} & train     & test       & 1273 & medical 4-way MC \\
WikiQA      & \texttt{wiki\_qa}                & train          & test (sub.)& 1500 & open-domain answer ranking \\
Winogrande  & \texttt{allenai/winogrande (xl)} & train          & validation & 1267 & pronoun-resolution 2-way \\
\bottomrule
\end{tabular}%
}
\end{table}

\subsection{Held-out benchmarks}
\label{app:setup-benchmarks}

Tab.~\ref{tab:benchmarks} lists the nine held-out cross-domain
benchmarks evaluated identically across every (model, task, method)
combination, together with their HuggingFace dataset identifiers,
splits, sample sizes, shot counts, scoring metrics, and
Knowledge / Cognition cohort assignment. Five benchmarks (ARC-Challenge,
GSM8K, IFEval, Math-500, TruthfulQA) use their full eval splits;
HellaSwag, TriviaQA, and MNLI are subsetted as noted. MMLU uses a
stratified subset of $30$ questions per subject across all $57$
subjects, totalling $1710$ items, so each subject contributes equal
weight regardless of its original size.

\begin{table}[h]
\centering
\caption{The nine held-out cross-domain benchmarks evaluated
identically across every (model, task, method) combination. The
IFEval metric is the average of prompt $\times$ instruction strict
$\times$ loose accuracy.}
\label{tab:benchmarks}
\scriptsize
\setlength{\tabcolsep}{4pt}
\resizebox{\textwidth}{!}{%
\begin{tabular}{@{}llllrll@{}}
\toprule
Benchmark & HF dataset & Eval split & Eval $n$ & $n$-shot & Metric & Dimension \\
\midrule
ARC-Challenge & \texttt{allenai/ai2\_arc (ARC-Challenge)}      & test                       & 1172 & 0          & accuracy        & Cognition \\
GSM8K         & \texttt{gsm8k (main)}                          & test                       & 1319 & 8-shot CoT & exact-number    & Cognition \\
HellaSwag     & \texttt{Rowan/hellaswag}                       & validation (sub.)          & 1500 & 0          & accuracy        & Cognition \\
IFEval        & \texttt{google/IFEval}                         & train                      &  541 & 0          & see caption     & Cognition \\
Math-500      & \texttt{HuggingFaceH4/MATH-500}                & test                       &  500 & 0-shot CoT & accuracy (boxed)& Cognition \\
MMLU          & \texttt{cais/mmlu (all)}                       & test (stratified)          & 1710 & 5-shot     & accuracy        & Knowledge \\
MNLI          & \texttt{nyu-mll/multi\_nli}                    & validation\_matched (sub.) & 2000 & 0          & accuracy        & Cognition \\
TriviaQA      & \texttt{mandarjoshi/trivia\_qa (rc.nocontext)} & validation (sub.)          & 1500 & 0          & exact-match     & Knowledge \\
TruthfulQA    & \texttt{truthful\_qa (multiple\_choice)}       & validation                 &  817 & 0          & MC1 accuracy    & Knowledge \\
\bottomrule
\end{tabular}%
}
\end{table}

\subsection{Safety benchmarks and judges}
\label{app:safety-eval}

The safety axes reported in Tab.~\ref{tab:safety-harmbench} of the
main paper are evaluated identically across every (model, task,
method) cell, with no alignment data entering the repair procedure.
Generation uses the same policy as all other benchmarks: vLLM with
no \texttt{max\_tokens} cap (capping would bias refusal length on
either side), greedy decoding, and the model's default chat
template; the prompt set and judge for each axis are summarized in
Tab.~\ref{tab:safety-eval}.

\begin{table}[h]
\centering
\caption{Safety benchmark configurations.
HarmBench refusal is reported as the share of harmful prompts the
model declines (higher is safer). XSTest over-refusal is reported on
the safe half of v2 only (the ``looks unsafe but isn't'' subset);
lower is better. StrongREJECT reports the rubric harmfulness score
on a $0$-$1$ scale aggregated as the official convex combination of
refusal, specificity, and convincingness; lower is better.}
\label{tab:safety-eval}
\small
\setlength{\tabcolsep}{4pt}
\resizebox{\textwidth}{!}{%
\begin{tabular}{@{}llrll@{}}
\toprule
Benchmark & HF dataset & Eval $n$ & Judge & Aggregate \\
\midrule
HarmBench v1.0       & \texttt{walledai/HarmBench}         & 200 & GPT-4o classifier                          & refusal rate (\%) \\
XSTest v2 (safe)     & \texttt{natolambert/xstest-v2-copy} & 250 & GPT-4o 3-class~\citep{rottger2024xstest}   & over-refusal rate (\%) \\
StrongREJECT v1.0    & \texttt{walledai/StrongREJECT}      & 313 & GPT-4o-mini rubric~\citep{souly2024strongreject} & harmfulness $\in [0,1]$ \\
\bottomrule
\end{tabular}%
}
\end{table}

The HarmBench classifier follows the protocol of
\citet{mazeika2024harmbench}: each (prompt, response) pair is
classified as ``refusal'' or ``harmful compliance'' and the reported
score is the refusal share. XSTest uses the safe half of v2
($250$ prompts the model should comply with despite surface cues);
the judge follows the three-class protocol of
\citet{rottger2024xstest} and we count any non-full-compliance
(partial or refusal) as over-refusal. StrongREJECT applies its
$0$-$1$ rubric verbatim to every (prompt, response) pair; harmfulness
is averaged over prompts. No category- or severity-level reweighting
is applied; all benchmarks use unweighted means over their prompt
sets.

\subsection{Training hyperparameters}
\label{app:setup-hparams}

Tab.~\ref{tab:hparams} lists the uniform training configuration used
for every (model, task) cell, so any difference in repair outcome is
attributable to the repair method rather than to training-time
variance.

\begin{table}[h]
\centering
\caption{Uniform training hyperparameters used for every (model,
task) cell.}
\label{tab:hparams}
\small
\begin{tabular}{@{}lll@{}}
\toprule
Hyperparameter & Value & Notes \\
\midrule
Optimiser              & AdamW                                  & $\beta_1 = 0.9$, $\beta_2 = 0.999$, $\varepsilon = 10^{-8}$ \\
Peak learning rate     & $1.0 \times 10^{-5}$                   & reached at end of warm-up \\
Warm-up                & $3\%$ of total steps                   & linear from $0$ to peak \\
LR schedule            & linear decay to $0$                    & over remaining $97\%$ of steps \\
Weight decay           & $0$                                    & none \\
Per-device batch size  & $2$                                    & constrained by single A100-80GB \\
Gradient accumulation  & $32$                                   & effective batch $= 64$ \\
Effective batch size   & $64$                                   & per\_device $\times$ grad\_accum \\
Epochs                 & $3$                                    & \\
Max sequence length    & $4096$ tokens                          & task-specific within this ceiling \\
Precision              & \texttt{bfloat16} / \texttt{fp32} Adam & \\
Gradient checkpointing & on for full-parameter methods          & off for LoRA \\
Random seed            & $42$                                   & single seed per cell \\
\bottomrule
\end{tabular}
\end{table}

\subsection{Repair scope}
\label{app:repair-scope}

The post-hoc methods we evaluate (DG-Hard, WiSE-FT, FAPM; with
TIES and DARE-Linear additionally evaluated in the
App.~\ref{app:abl-merging} ablation) are applied per-tensor to
every weight matrix with
$\mathrm{ndim} \ge 2$ and at least $1024$ elements; tensors with
$\mathrm{ndim} > 2$ are reshaped to a $2$D matrix before SVD.
One-dimensional parameters (biases, RMSNorm scales) are left at
their fine-tuned values, since they do not admit a non-trivial
Marchenko-Pastur edge. On Llama-3.2-3B-Instruct this scope covers
every attention and MLP projection of the text decoder, plus
\texttt{embed\_tokens} and \texttt{lm\_head}. On Qwen3.5-4B the same
filter additionally accepts the vision-encoder weights and the
multi-token-prediction head; because inference is text-only, these
tensors are stripped at vLLM load
(\texttt{checkpoint\_compat.py:\_DROP\_PREFIXES}), and any
modifications computed on them during repair never reach evaluation.
The state-space \texttt{linear\_attn} blocks contribute
one-dimensional parameters (\texttt{A\_log}, \texttt{dt\_bias}) that
are skipped under the dimensionality rule; their
\texttt{conv1d.weight} is three-dimensional and is therefore reshaped
and repaired like any other matrix. The training-time methods
(L1-reg, V-SoftMask, CoFi-Tune, LoRA) see the model in their
respective canonical scopes during fine-tuning and inherit no
scope-restriction at inference time.

\section{Per-cell forgetting and balance scores}
\label{app:per-cell-tables}

This appendix holds the two per-cell tables that the body cites in
\S\ref{sec:per-cell-forgetting} (also labeled
\S\ref{sec:per-cell-balance}).
Tab.~\ref{tab:cell-damage} reports the (model, task) cell-level
fine-tuning impact: how many of the nine held-out benchmarks were
damaged ($\Delta_{\mathrm{FT}} \le -3$\,pp), how many were improved
($\Delta_{\mathrm{FT}} \ge +3$\,pp), the average and worst drops over
the damaged set, and the on-task gain
$\Delta_{\mathrm{FT, task}}$ (Full-SFT minus base). The remaining
($9 - \text{damaged} - \text{improved}$) benchmarks per cell are
unchanged within $\pm 3$\,pp.
Tab.~\ref{tab:cell-balance} reports the per-cell balance score, the
harmonic mean of (i) the method's mean held-out ratio against base
and (ii) the method's on-task ratio against Full-SFT, each multiplied
by $100$. The per-cell view uses these mean ratios because the
partition-conditional sub-statistics (\% healed, \% preserved) become
unstable on the small per-cell $n \approx 9$. FAPM is omitted from
the balance table because it never wins: its $90\%$ entry reversion
drives the held-out ratio toward base but collapses the on-task
ratio, leaving it well behind both DG-Hard and WiSE-FT in every cell.

\begin{table}[h]
\centering
\caption{Per-cell impact of fine-tuning across the $14$ (model, task)
cells of our experimental matrix. Fine-tuning damages at least one
held-out benchmark in $13$ of $14$ cells and incidentally improves
more held-out benchmarks than it damages ($55$ vs $30$ in total).}
\label{tab:cell-damage}
\small
\setlength{\tabcolsep}{4pt}
\resizebox{\linewidth}{!}{%
\begin{tabular}{llccrlr}
\toprule
Model & Task & \# damaged $\downarrow$ & \# improved & Avg drop on damaged (pp) & Worst single drop (pp) & $\Delta_{\mathrm{FT,task}}$ (pp) $\uparrow$ \\
\midrule
Qwen  & RTE        & 2 & 5 & $-13.48$ & 17.7 (TriviaQA)   & $+10.1$ \\
Qwen  & StrategyQA & 2 & 5 & $-4.84$  & 6.2 (Math)        & $+5.2$  \\
Qwen  & ReClor     & 3 & 5 & $-9.95$  & 17.4 (Math)       & $+37.2$ \\
Qwen  & BoolQ      & 2 & 5 & $-6.02$  & 8.4 (Math)        & $+13.3$ \\
Qwen  & MedQA      & \textbf{4} & 4 & $\bm{-33.47}$ & \textbf{91.8 (GSM8K)} & $+12.2$ \\
Qwen  & WikiQA     & 2 & 5 & $-11.67$ & 11.8 (TriviaQA)   & $+46.8$ \\
Qwen  & Winogrande & 3 & 3 & $\bm{-25.61}$ & 35.9 (GSM8K) & $+25.8$ \\
Llama & RTE        & 3 & 3 & $-7.34$  & 9.1 (GSM8K)       & $+31.0$ \\
Llama & StrategyQA & 3 & 3 & $-10.36$ & 17.0 (GSM8K)      & $+7.0$  \\
Llama & ReClor     & 0 & 4 & (none)   & (none; max $-2.3$ IFEval) & $+19.4$ \\
Llama & BoolQ      & 2 & 2 & $-4.69$  & 6.0 (IFEval)      & $+10.4$ \\
Llama & MedQA      & 1 & 4 & $-3.60$  & 3.6 (MNLI)        & $+8.2$  \\
Llama & WikiQA     & 2 & 3 & $-5.64$  & 7.0 (GSM8K)       & $+1.4$  \\
Llama & Winogrande & 1 & 4 & $-7.71$  & 7.7 (TruthfulQA)  & $+32.3$ \\
\midrule
\textbf{Total} & (across 14 cells) & \textbf{30} & \textbf{55} & (mean over 30 damaged: $-13.3$) & (max $91.8$) & (avg $+18.6$) \\
\bottomrule
\end{tabular}%
}
\end{table}

\begin{table}[h]
\centering
\caption{Per-cell balance scores per (model, task) cell, plus the
cohort aggregation of the winner column. The balance score is the
harmonic mean of the method's mean held-out ratio (vs base) and on-task
ratio (vs FT), each multiplied by $100$. FAPM~\citep{huang2025fapm} is excluded because it
never wins on balance.}
\label{tab:cell-balance}
\small
\begin{tabular}{llccl}
\toprule
Model & Task & bal(WiSE) $\uparrow$ & bal(DG) $\uparrow$ & Winner \\
\midrule
Qwen  & RTE        & $103.6_{\pm 0.4}$ & $\bm{104.3_{\pm 0.4}}$ & DG-Hard \\
Qwen  & StrategyQA & $102.7_{\pm 0.6}$ & $\bm{104.6_{\pm 0.6}}$ & DG-Hard \\
Qwen  & ReClor     & $98.6_{\pm 0.4}$  & $\bm{101.5_{\pm 0.5}}$ & DG-Hard \\
Qwen  & BoolQ      & $105.9_{\pm 0.3}$ & $\bm{108.5_{\pm 0.3}}$ & DG-Hard \\
Qwen  & MedQA      & $105.4_{\pm 0.5}$ & $\bm{105.4_{\pm 0.5}}$ & tie: DG-Hard $\approx$ WiSE-FT \\
Qwen  & WikiQA     & $101.9_{\pm 0.2}$ & $\bm{103.4_{\pm 0.3}}$ & DG-Hard \\
Qwen  & Winogrande & $98.9_{\pm 0.3}$  & $\bm{102.8_{\pm 0.3}}$ & DG-Hard \\
Llama & RTE        & $\bm{100.0_{\pm 0.6}}$ & $98.8_{\pm 0.6}$  & WiSE-FT \\
Llama & StrategyQA & $\bm{101.1_{\pm 0.7}}$ & $99.2_{\pm 0.7}$  & WiSE-FT \\
Llama & ReClor     & $\bm{97.4_{\pm 0.9}}$  & $96.0_{\pm 0.8}$  & WiSE-FT \\
Llama & BoolQ      & $\bm{100.7_{\pm 0.3}}$ & $100.2_{\pm 0.3}$ & WiSE-FT \\
Llama & MedQA      & $102.1_{\pm 0.6}$ & $101.9_{\pm 0.6}$ & tie: DG-Hard $\approx$ WiSE-FT \\
Llama & WikiQA     & $100.8_{\pm 0.3}$ & $101.3_{\pm 0.2}$ & tie: DG $\approx$ WiSE-FT \\
Llama & Winogrande & $97.4_{\pm 0.4}$  & $\bm{99.5_{\pm 0.4}}$  & DG-Hard \\
\midrule
\multicolumn{5}{l}{\emph{Aggregating the winner column:}} \\
\midrule
\multicolumn{2}{l}{Cohort}                       & DG-Hard wins & WiSE-FT wins & Ties \\
\multicolumn{2}{l}{Qwen (reasoning, $n=7$)}      & \textbf{6}   & 0            & 1 \\
\multicolumn{2}{l}{Llama (non-reasoning, $n=7$)} & 1            & \textbf{4}   & 2 \\
\multicolumn{2}{l}{All cells ($n=14$)}           & 7            & 4            & 3 \\
\bottomrule
\end{tabular}
\end{table}

\section{Per-cohort Clean-up and Retention}
\label{app:cohort-cleanup-retention}

Tabs.~\ref{tab:cohort-cleanup} and~\ref{tab:cohort-retention} report
the per-cohort Clean-up and Retention sub-scores for every method,
computed exactly as in \S\ref{sec:recovery-preservation}: Clean-up
$= \mathrm{HM}(\text{\% healed},\,\text{Non-damage rate})$ and
Retention $= \mathrm{HM}(\text{\% preserved},\,\text{On-task
retention})$, with the partition-conditional sub-statistics
themselves clipped to $[0, 100]$ before the harmonic mean. The
Knowledge and Cognition rows are the source of the per-cohort
numbers cited in \S\ref{sec:knowledge-cognition}; the Overall row
matches the population-level Tab.~\ref{tab:headline-method}. A
$0$ entry in the Retention table indicates that the underlying
\% preserved averaged non-positive on the improved partition, so the
clipped HM collapses to zero.

\begin{table}[h]
\centering
\caption{Per-cohort Clean-up $= \mathrm{HM}(\text{\% healed},\,
\text{Non-damage rate})$, in $[0, 100]$, higher is better; $> 100$
appears when \% healed overshoots base on the damaged partition.
\textbf{Bold} = best per cohort; \underline{underline} = second-best.}
\label{tab:cohort-cleanup}
\small
\setlength{\tabcolsep}{6pt}
\begin{tabular}{lrrrrr}
\toprule
Method & Overall & Knowledge & Cognition & Reasoning & Non-reasoning \\
\midrule
DG-Hard    &  $83.1_{\pm 2.8}$ &  $85.3_{\pm 3.2}$ & $\underline{81.6_{\pm 3.5}}$ &  $73.4_{\pm 2.6}$ & $\underline{92.9_{\pm 4.3}}$ \\
WiSE-FT    &  $82.7_{\pm 2.7}$ &  $87.5_{\pm 3.7}$ &  $79.6_{\pm 3.2}$ &  $80.3_{\pm 2.3}$ &  $83.6_{\pm 4.8}$ \\
L1-reg     &  $83.4_{\pm 2.6}$ & $\bm{102.0_{\pm 2.7}}$ &  $79.5_{\pm 3.1}$ & $\underline{92.1_{\pm 2.3}}$ &  $75.7_{\pm 4.2}$ \\
CoFi-Tune  &  $77.8_{\pm 2.7}$ &  $97.1_{\pm 2.1}$ &  $73.7_{\pm 3.1}$ & $\bm{92.3_{\pm 2.1}}$ &  $67.8_{\pm 4.5}$ \\
V-SoftMask &  $24.9_{\pm 5.9}$ &  $51.9_{\pm 3.4}$ &  $18.3_{\pm 5.7}$ &  $42.2_{\pm 3.5}$ &   $0.0_{\pm 3.5}$ \\
LoRA       & $\underline{85.6_{\pm 2.7}}$ &  $80.5_{\pm 2.3}$ & $\underline{86.5_{\pm 3.2}}$ &  $80.0_{\pm 2.5}$ &  $89.4_{\pm 4.4}$ \\
FAPM       & $\bm{91.9_{\pm 2.7}}$ & $\underline{98.1_{\pm 3.9}}$ & $\bm{90.6_{\pm 3.1}}$ &  $87.9_{\pm 2.7}$ & $\bm{95.5_{\pm 4.1}}$ \\
\bottomrule
\end{tabular}
\end{table}

\begin{table}[h]
\centering
\caption{Per-cohort Retention $= \mathrm{HM}(\text{\% preserved},\,
\text{On-task retention})$, in $[0, 100]$, higher is better. Entries
of $0$ indicate that the method's average \% preserved on the
improved partition was non-positive, collapsing the harmonic mean
to zero. \textbf{Bold} = best per cohort; \underline{underline} =
second-best.}
\label{tab:cohort-retention}
\small
\setlength{\tabcolsep}{6pt}
\begin{tabular}{lrrrrr}
\toprule
Method & Overall & Knowledge & Cognition & Reasoning & Non-reasoning \\
\midrule
DG-Hard    & $\underline{83.5_{\pm 1.6}}$ & $\underline{84.3_{\pm 1.7}}$ & $\underline{82.5_{\pm 2.4}}$ & $\underline{89.1_{\pm 1.6}}$ & $75.3_{\pm 2.5}$ \\
WiSE-FT    & $71.0_{\pm 2.2}$ & $80.3_{\pm 1.9}$ & $57.5_{\pm 3.9}$ & $66.4_{\pm 2.7}$ & $\underline{76.8_{\pm 2.2}}$ \\
L1-reg     & $55.6_{\pm 3.5}$ & $78.5_{\pm 2.2}$ & $13.0_{\pm 6.1}$ & $45.0_{\pm 5.5}$ & $68.1_{\pm 3.5}$ \\
CoFi-Tune  & $28.5_{\pm 6.2}$ & $58.9_{\pm 2.8}$ & $0.0_{\pm 0.2}$ & $5.2_{\pm 3.9}$ & $52.4_{\pm 4.8}$ \\
V-SoftMask & $\bm{97.1_{\pm 1.2}}$ & $\bm{97.2_{\pm 1.3}}$ & $\bm{97.0_{\pm 1.4}}$ & $\bm{95.5_{\pm 1.1}}$ & $\bm{99.2_{\pm 1.6}}$ \\
LoRA       & $9.0_{\pm 3.5}$ & $40.5_{\pm 3.7}$ & $0.0_{\pm 0.0}$ & $0.0_{\pm 1.1}$ & $29.0_{\pm 6.2}$ \\
FAPM       & $0.0_{\pm 0.3}$ & $0.0_{\pm 2.7}$ & $0.0_{\pm 0.0}$ & $0.0_{\pm 0.0}$ & $0.0_{\pm 3.1}$ \\
\bottomrule
\end{tabular}
\end{table}

\paragraph{Per-axis trends.}
The two tables make the structural asymmetry of the baseline field
explicit. On Retention (Tab.~\ref{tab:cohort-retention}),
\textbf{V-SoftMask} dominates every cohort by a wide margin, scoring
$\geq 95$ on all five and exceeding the next-best method by $14$ to
$28$\,pp on each. This is consistent with V-SoftMask's mechanism: by
damping gradient flow during fine-tuning it leaves the held-out and
on-task distributions closer to the base, so on the improved
partition it inherits the same scores it produced and thus high
\% preserved. The same mechanism explains its position on Clean-up
(Tab.~\ref{tab:cohort-cleanup}), where V-SoftMask is the
\emph{lowest}-scoring method on Overall ($24.9$) and Cognition
($18.3$) and falls to $0$ on Non-reasoning. Because gradient damping
suppresses both the noise residual and any genuine recovery, the
method has nothing with which to bring damaged held-out benchmarks
back toward base, so its Clean-up component collapses. The mirror
extremes are \textbf{FAPM} and \textbf{LoRA}, which post some of the
highest Clean-up scores ($91.9$ and $85.6$ Overall, respectively)
yet hit $0$ on every Cognition and Reasoning Retention cell because
their average \% preserved on the improved partition is non-positive
(FAPM's $90\%$ sparsity reversion drags every benchmark toward base;
LoRA's adapter delta never produced the improved-partition lifts in
the first place). \textbf{L1-reg} and \textbf{CoFi-Tune} are partial
versions of the same single-axis pattern: high Clean-up, weak or
zero Retention on Cognition.

\paragraph{Why balance is required.}
Combined $= \mathrm{HM}(\text{Clean-up},\,\text{Retention})$ is
designed precisely to demote single-axis extremes via the
bottleneck property of the harmonic mean: any method with a near-zero
component cannot recover at the headline level regardless of how
strong its other axis is. The two tables in this section make the
mechanism visible cohort by cohort. V-SoftMask's perfect Retention
is undone by its near-zero Clean-up; FAPM's strong Clean-up is
undone by its identically-zero Retention; LoRA and CoFi-Tune are
similarly imbalanced in less extreme but still disqualifying ways.
DG-Hard is the only method that scores in the upper half of both
axes simultaneously across all five cohorts, which is consistent
with its leading the population-level Combined metric in
Tab.~\ref{tab:headline-method}. At the cohort level, DG-Hard is not
required to win Combined everywhere (\S\ref{sec:knowledge-cognition}
notes that L1-reg edges it on Knowledge): the structural claim is
that DG-Hard is the only method that does not collapse on either
axis in any cohort, not that it wins Combined in every cohort.

\section{Full per-cohort sub-score breakdown}
\label{app:cohort-full}

\S\ref{sec:two-views} summarized the per-cohort
Combined scores for DG-Hard vs WiSE-FT. The full Clean-up and
Retention sub-scores per cohort (the harmonic-mean inputs to Combined)
are tabulated here. DG-Hard's edge over WiSE-FT comes mostly from
Retention: their Clean-up scores are within $\sim 3$\,pp on most
cohorts, but DG-Hard's Retention is $12{+}$\,pp higher on Overall,
Cognition, and Reasoning, driven mainly by held-out preservation
since on-task retention is essentially flat at $\sim 97\%$ for both
methods.

\begin{table}[h]
\centering
\caption{Per-cohort Clean-up / Retention / Combined for DG-Hard vs
WiSE-FT~\citep{wortsman2022robust}. \textbf{Bold} marks the higher Combined score in this
\emph{head-to-head} DG-Hard vs WiSE-FT comparison only; it is
\emph{not} a cross-method best-per-cohort indicator. For
cross-method per-cohort winners across all baselines, see
App.~\ref{app:cohort-cleanup-retention}
(Tabs.~\ref{tab:cohort-cleanup} and~\ref{tab:cohort-retention}); per
\S\ref{sec:knowledge-cognition}, DG-Hard wins Combined on four of
five cohorts at the cross-method level (L1-reg edges it on
Knowledge).}
\label{tab:cohort-full}
\small
\setlength{\tabcolsep}{5pt}
\resizebox{\textwidth}{!}{%
\begin{tabular}{lrrrrrr}
\toprule
Cohort & DG Clean-up & DG Retention & DG Combined $\uparrow$ & WiSE Clean-up & WiSE Retention & WiSE Combined $\uparrow$ \\
\midrule
Overall                & $83.1_{\pm 2.8}$ & $83.5_{\pm 1.6}$ & \textbf{83.3} & $82.7_{\pm 2.7}$ & $71.0_{\pm 2.2}$ & 76.4 \\
Knowledge              & $85.3_{\pm 3.2}$ & $84.3_{\pm 1.7}$ & \textbf{84.8} & $87.5_{\pm 3.7}$ & $80.3_{\pm 1.9}$ & 83.7 \\
Cognition              & $81.6_{\pm 3.5}$ & $82.5_{\pm 2.4}$ & \textbf{82.1} & $79.6_{\pm 3.2}$ & $57.5_{\pm 3.9}$ & 66.8 \\
Reasoning (Qwen)       & $73.4_{\pm 2.6}$ & $89.1_{\pm 1.6}$ & \textbf{80.5} & $80.3_{\pm 2.3}$ & $66.4_{\pm 2.7}$ & 72.7 \\
Non-reasoning (Llama)  & $92.9_{\pm 4.3}$ & $75.3_{\pm 2.5}$ & \textbf{83.2} & $83.6_{\pm 4.8}$ & $76.8_{\pm 2.2}$ & 80.1 \\
\bottomrule
\end{tabular}%
}
\end{table}

\section{Per-(model, task, method, benchmark) results}
\label{app:percell-bench}

For full transparency, Tab.~\ref{tab:bench-qwen} and
Tab.~\ref{tab:bench-llama} list every per-cell, per-benchmark score in
the experimental matrix ($1134$ measurements in total: $2$ models
$\times\ 7$ tasks $\times\ 9$ method rows $\times\ 9$ held-out
benchmarks; each table additionally reports $3$ derived columns
(\emph{Avg.}, \emph{Results}, \emph{Combined}) per row).
Each task block contains nine method rows (Pre-trained, Full-SFT, six
repair baselines, plus our DG-Hard); within each task block,
\textbf{bold} marks the best score per column and \underline{underline}
marks the second-best (when separated by $\geq 0.001$). The right-most
three columns are \emph{Avg.} (mean of the nine held-out benchmarks),
\emph{Results} (on-task \texttt{task\_\{name\}} score), and
\emph{Combined} (defined in the table caption).

\begin{table}[!htbp]
\centering
\caption{Per-benchmark held-out scores for Qwen3.5-4B across all
(task, method) cells. \textbf{Bold} = best per column within the task
block; \underline{underline} = second-best (gap $\geq 0.001$ from
best). Avg.\ = mean of the nine held-out benchmarks. Results = on-task
\texttt{task\_\{name\}} score. Combined $= \mathrm{HM}(\overline{\text{\%
healed}},\, \overline{\text{\% preserved}})$ on the partitioned
held-out set, with per-benchmark \% healed and \% preserved clipped to
$[0, 100]$ before averaging; cells with no damaged or no improved
benchmarks default the corresponding side to $100$. Pre-trained and
Full-SFT have no defined Combined (they parameterize the partition).
Bold/underline on Combined follow the same convention as other columns,
restricted to the seven repair methods (excluding the two reference
rows). ``Repair methods'' here covers both training-time interventions
(L1-reg, V-SoftMask~\citep{ke2023continual}, CoFi-Tune~\citep{zhang2024cofitune}, LoRA~\citep{hu2022lora}) and post-hoc methods (WiSE-FT~\citep{wortsman2022robust},
FAPM~\citep{huang2025fapm}, DG-Hard); see \S\ref{sec:related-work} for the full bucketing.}
\label{tab:bench-qwen}
\resizebox{\textwidth}{!}{%
\footnotesize
\begin{tabular}{llcccccccccccc}
\toprule
Task & Method & ARC & GSM8K & HSwag & IFEval & Math & MMLU & MNLI & Trivia & TruthQA & Avg. & Results & Combined \\
\midrule
\multirow{9}{*}{MedQA}
 & Pre-trained   & 0.8643 & 0.9295 & 0.5060 & 0.3460 & 0.5320 & 0.6041 & \textbf{0.8464} & 0.4767 & 0.6548 & 0.6400 & 0.6308 & --- \\
 & Full-SFT      & \textbf{0.9138} & 0.0114 & 0.8780 & 0.3026 & 0.2640 & \textbf{0.7550} & 0.8270 & 0.3673 & 0.7552 & 0.5638 & 0.7526 & --- \\
 & L1-reg        & 0.6374 & \textbf{0.9401} & 0.4433 & 0.3504 & \underline{0.5920} & 0.6526 & \underline{0.8425} & 0.5020 & 0.6304 & 0.6212 & 0.7494 & 0.1489 \\
 & WiSE-FT       & 0.8993 & 0.8590 & \underline{0.8787} & 0.3551 & 0.5920 & 0.7444 & 0.8185 & 0.4967 & \underline{0.7625} & \textbf{0.7118} & 0.7541 & \underline{0.9437} \\
 & V-SoftMask    & \underline{0.9078} & 0.1099 & 0.8760 & 0.3140 & 0.2520 & \underline{0.7526} & 0.8265 & 0.3887 & 0.7491 & 0.5752 & 0.7423 & 0.2459 \\
 & CoFi-Tune     & 0.7799 & \underline{0.9348} & 0.4667 & \textbf{0.3554} & 0.5500 & 0.6240 & 0.8365 & 0.4833 & 0.6622 & 0.6325 & 0.7494 & 0.0975 \\
 & LoRA          & 0.7526 & 0.9022 & 0.5413 & 0.3514 & 0.5160 & 0.6240 & 0.8115 & 0.4600 & 0.6255 & 0.6205 & 0.7321 & 0.1069 \\
 & FAPM          & 0.8686 & 0.9280 & 0.5013 & 0.3458 & 0.4980 & 0.6047 & 0.8370 & 0.4673 & 0.6438 & 0.6327 & 0.6088 & 0.0440 \\
 & DG-Hard (Ours)& 0.9019 & 0.8158 & \textbf{0.8820} & 0.3501 & \textbf{0.6000} & 0.7462 & 0.8120 & \textbf{0.5027} & \textbf{0.7797} & \underline{0.7100} & \textbf{0.7549} & \textbf{0.9466} \\
\midrule
\multirow{9}{*}{WikiQA}
 & Pre-trained   & 0.8643 & 0.9295 & 0.5060 & 0.3460 & 0.5320 & 0.6041 & \textbf{0.8464} & 0.4767 & 0.6548 & 0.6400 & 0.4973 & --- \\
 & Full-SFT      & \textbf{0.8985} & \underline{0.9424} & \underline{0.8527} & \textbf{0.6053} & 0.5340 & \textbf{0.7971} & 0.7310 & 0.3587 & \underline{0.6965} & \underline{0.7129} & 0.9651 & --- \\
 & L1-reg        & 0.8353 & 0.9280 & 0.4987 & 0.3496 & 0.5020 & 0.6965 & 0.8250 & 0.4867 & \textbf{0.7173} & 0.6488 & 0.9674 & 0.4492 \\
 & WiSE-FT       & 0.8072 & \textbf{0.9447} & 0.5880 & 0.3670 & \underline{0.5520} & 0.6930 & 0.8225 & \textbf{0.5293} & 0.6891 & 0.6659 & 0.9630 & \underline{0.4720} \\
 & V-SoftMask    & \underline{0.8805} & 0.9393 & \textbf{0.8620} & \underline{0.5772} & \textbf{0.5580} & 0.7971 & 0.7550 & 0.4087 & 0.6756 & \textbf{0.7171} & 0.9659 & 0.4486 \\
 & CoFi-Tune     & 0.8524 & 0.9386 & 0.5460 & 0.3526 & 0.5060 & 0.6684 & 0.8180 & \underline{0.5040} & 0.6573 & 0.6493 & 0.9648 & 0.1900 \\
 & LoRA          & 0.8652 & 0.9318 & 0.5167 & 0.3478 & 0.5240 & 0.6433 & 0.7905 & 0.4707 & 0.6769 & 0.6407 & \textbf{0.9679} & 0.2613 \\
 & FAPM          & 0.8652 & 0.9325 & 0.4820 & 0.3524 & 0.4860 & 0.6123 & \underline{0.8420} & 0.4693 & 0.6389 & 0.6312 & 0.4986 & 0.0360 \\
 & DG-Hard (Ours)& 0.7867 & 0.9424 & 0.7953 & 0.4555 & 0.5380 & 0.7316 & 0.7650 & 0.4893 & 0.6928 & 0.6885 & 0.9613 & \textbf{0.6039} \\
\midrule
\multirow{9}{*}{Winogrande}
 & Pre-trained   & \underline{0.8643} & \underline{0.9295} & 0.5060 & 0.3460 & 0.5320 & 0.6041 & \underline{0.8464} & 0.4767 & 0.6548 & 0.6400 & 0.6393 & --- \\
 & Full-SFT      & 0.6305 & 0.5709 & \underline{0.8727} & \underline{0.4166} & 0.5520 & 0.7520 & 0.8170 & 0.3007 & 0.6573 & 0.6189 & \underline{0.8974} & --- \\
 & L1-reg        & 0.8302 & \textbf{0.9386} & 0.4407 & 0.3511 & \textbf{0.5700} & 0.6690 & \textbf{0.8530} & \underline{0.4973} & \underline{0.6842} & 0.6482 & 0.8627 & 0.2888 \\
 & WiSE-FT       & 0.8063 & 0.9174 & 0.6027 & 0.3619 & \underline{0.5600} & 0.7058 & 0.8305 & 0.4900 & 0.6267 & 0.6557 & 0.8579 & 0.5473 \\
 & V-SoftMask    & 0.8336 & 0.8287 & \textbf{0.8753} & \textbf{0.5039} & 0.5600 & \textbf{0.7930} & 0.8250 & 0.4353 & 0.6450 & \textbf{0.7000} & \textbf{0.8998} & \textbf{0.8790} \\
 & CoFi-Tune     & 0.8183 & 0.9212 & 0.4620 & 0.3492 & 0.5200 & 0.6567 & 0.8435 & 0.4727 & \textbf{0.7001} & 0.6382 & 0.8477 & 0.2331 \\
 & LoRA          & 0.8038 & 0.9052 & 0.0613 & 0.3510 & 0.5120 & 0.6713 & 0.7200 & 0.4513 & 0.4945 & 0.5523 & 0.8548 & 0.2901 \\
 & FAPM          & \textbf{0.8763} & 0.9265 & 0.4807 & 0.3481 & 0.5340 & 0.6246 & 0.8455 & 0.4813 & 0.6463 & 0.6404 & 0.6385 & 0.1059 \\
 & DG-Hard (Ours)& 0.7415 & 0.8795 & 0.8527 & 0.3656 & 0.5500 & \underline{0.7661} & 0.8405 & \textbf{0.5053} & 0.6095 & \underline{0.6790} & 0.8950 & \underline{0.7592} \\
\midrule
\multirow{9}{*}{StrategyQA}
 & Pre-trained   & 0.8643 & \underline{0.9295} & 0.5060 & 0.3460 & \underline{0.5320} & 0.6041 & 0.8464 & 0.4767 & 0.6548 & 0.6400 & 0.7176 & --- \\
 & Full-SFT      & \textbf{0.8959} & 0.8946 & \textbf{0.8713} & \underline{0.6361} & 0.4700 & \textbf{0.7661} & 0.8278 & 0.4740 & \underline{0.7026} & \textbf{0.7265} & \textbf{0.7700} & --- \\
 & L1-reg        & 0.8874 & 0.9128 & 0.8500 & 0.5273 & 0.4840 & 0.7620 & 0.8279 & 0.4560 & 0.6830 & 0.7100 & \underline{0.7584} & 0.5037 \\
 & WiSE-FT       & 0.8951 & 0.9265 & 0.8353 & 0.3952 & 0.5180 & 0.7526 & 0.8323 & 0.4453 & 0.6659 & 0.6962 & 0.7496 & \textbf{0.7267} \\
 & V-SoftMask    & 0.8959 & 0.8939 & \underline{0.8653} & \textbf{0.6553} & 0.4560 & 0.7661 & 0.8258 & 0.4627 & 0.6891 & \underline{0.7233} & 0.7496 & 0.0000 \\
 & CoFi-Tune     & 0.8003 & \textbf{0.9393} & 0.5267 & 0.3691 & 0.4840 & 0.6199 & \textbf{0.8531} & \underline{0.4887} & \textbf{0.7087} & 0.6433 & 0.7555 & 0.3518 \\
 & LoRA          & 0.8635 & 0.9287 & 0.5027 & 0.3470 & \textbf{0.5460} & 0.6205 & \underline{0.8486} & \textbf{0.4933} & 0.6622 & 0.6458 & 0.6652 & 0.0982 \\
 & FAPM          & 0.8823 & 0.9265 & 0.4873 & 0.2106 & 0.3800 & 0.6287 & 0.8454 & 0.4600 & 0.6512 & 0.6080 & 0.7132 & 0.2188 \\
 & DG-Hard (Ours)& 0.8942 & 0.9105 & 0.8547 & 0.6003 & 0.4900 & 0.7632 & 0.8340 & 0.4600 & 0.6928 & 0.7222 & 0.7511 & \underline{0.5457} \\
\midrule
\multirow{9}{*}{BoolQ}
 & Pre-trained   & 0.8643 & 0.9295 & 0.5060 & 0.3460 & 0.5320 & 0.6041 & 0.8464 & 0.4767 & 0.6548 & 0.6400 & 0.7847 & --- \\
 & Full-SFT      & \underline{0.9070} & 0.9121 & \underline{0.8807} & \textbf{0.7440} & 0.4480 & 0.7731 & 0.8099 & 0.4760 & \underline{0.7809} & 0.7480 & 0.9177 & --- \\
 & L1-reg        & 0.7722 & \textbf{0.9477} & 0.5087 & 0.3722 & \underline{0.5780} & 0.7275 & 0.8399 & \textbf{0.5320} & 0.7748 & 0.6726 & 0.9180 & 0.5068 \\
 & WiSE-FT       & 0.7944 & \underline{0.9401} & 0.8653 & 0.4914 & 0.5500 & 0.7620 & 0.8279 & \underline{0.5080} & 0.7405 & 0.7200 & 0.9180 & \textbf{0.6579} \\
 & V-SoftMask    & \textbf{0.9121} & 0.9121 & \textbf{0.8853} & \underline{0.7194} & 0.4620 & \underline{0.7754} & 0.8066 & 0.4747 & \textbf{0.7846} & \underline{0.7480} & 0.9165 & 0.1537 \\
 & CoFi-Tune     & 0.7910 & 0.9249 & 0.5180 & 0.3618 & 0.5120 & 0.6585 & 0.8420 & 0.5067 & 0.7234 & 0.6487 & 0.9116 & 0.3052 \\
 & LoRA          & 0.8635 & 0.9378 & 0.5033 & 0.3412 & 0.5420 & 0.6327 & 0.8435 & 0.5073 & 0.6805 & 0.6502 & 0.9080 & 0.1386 \\
 & FAPM          & 0.8771 & 0.9303 & 0.4913 & 0.3514 & 0.5260 & 0.6094 & \textbf{0.8465} & 0.4640 & 0.6671 & 0.6403 & 0.7881 & 0.1618 \\
 & DG-Hard (Ours)& 0.9053 & 0.9249 & 0.8793 & 0.6475 & \textbf{0.6180} & \textbf{0.8053} & 0.8070 & 0.4640 & 0.7687 & \textbf{0.7578} & \textbf{0.9187} & \underline{0.6487} \\
\midrule
\multirow{9}{*}{ReClor}
 & Pre-trained   & 0.8643 & 0.9295 & 0.5060 & 0.3460 & \underline{0.5320} & 0.6041 & \textbf{0.8464} & 0.4767 & 0.6548 & 0.6400 & 0.5120 & --- \\
 & Full-SFT      & \textbf{0.9121} & 0.8529 & 0.8740 & 0.4442 & 0.3580 & \underline{0.7626} & 0.7986 & 0.4593 & 0.7234 & 0.6872 & \underline{0.8840} & --- \\
 & L1-reg        & 0.9087 & 0.9234 & 0.8773 & 0.3682 & 0.4080 & 0.7620 & 0.8193 & \underline{0.4893} & \underline{0.7442} & \textbf{0.7000} & 0.8660 & \textbf{0.6595} \\
 & WiSE-FT       & 0.9087 & 0.7506 & \textbf{0.8780} & 0.3086 & 0.3920 & 0.7556 & 0.8164 & 0.2587 & 0.7185 & 0.6430 & 0.8560 & 0.3036 \\
 & V-SoftMask    & 0.9053 & 0.8696 & 0.8707 & \underline{0.4445} & 0.3420 & \textbf{0.7661} & 0.7955 & 0.4733 & 0.7271 & \underline{0.6882} & \textbf{0.8900} & 0.1351 \\
 & CoFi-Tune     & 0.9061 & \textbf{0.9363} & 0.8240 & 0.2495 & 0.3860 & 0.6930 & 0.7948 & \textbf{0.5047} & \textbf{0.7601} & 0.6727 & 0.8200 & 0.4879 \\
 & LoRA          & 0.8251 & 0.9037 & 0.8220 & 0.2193 & 0.3780 & 0.6807 & 0.8007 & 0.4780 & 0.7393 & 0.6497 & 0.8080 & 0.3460 \\
 & FAPM          & 0.8677 & 0.9363 & 0.4993 & 0.3553 & \textbf{0.5460} & 0.6018 & \underline{0.8385} & 0.4707 & 0.6499 & 0.6406 & 0.4860 & 0.0640 \\
 & DG-Hard (Ours)& \underline{0.9096} & 0.8249 & 0.8767 & \textbf{0.4543} & 0.5180 & 0.7591 & 0.8085 & 0.2887 & 0.7283 & 0.6853 & 0.8520 & \underline{0.5440} \\
\midrule
\multirow{9}{*}{RTE}
 & Pre-trained   & \underline{0.8643} & 0.9295 & 0.5060 & 0.3460 & 0.5320 & 0.6041 & 0.8464 & 0.4767 & 0.6548 & 0.6400 & 0.8484 & --- \\
 & Full-SFT      & 0.7713 & 0.9242 & \textbf{0.8847} & \underline{0.4150} & \textbf{0.6140} & \textbf{0.7901} & 0.8202 & 0.3000 & 0.7356 & 0.6950 & \textbf{0.9495} & --- \\
 & L1-reg        & 0.7841 & 0.9348 & 0.8773 & \textbf{0.4327} & \underline{0.5980} & 0.7538 & \textbf{0.8508} & 0.4780 & 0.7173 & \textbf{0.7141} & \underline{0.9386} & \underline{0.6887} \\
 & WiSE-FT       & 0.8157 & \underline{0.9409} & 0.7773 & 0.3883 & 0.5640 & 0.7082 & 0.8455 & \underline{0.4927} & 0.7356 & 0.6965 & 0.9386 & \textbf{0.6948} \\
 & V-SoftMask    & 0.8635 & 0.9348 & 0.7847 & 0.3547 & 0.5340 & 0.6766 & 0.8500 & 0.3767 & \textbf{0.7491} & 0.6804 & 0.9314 & 0.5554 \\
 & CoFi-Tune     & 0.8456 & \textbf{0.9439} & 0.5187 & 0.3488 & 0.4780 & 0.6889 & 0.8415 & \textbf{0.5153} & 0.7136 & 0.6549 & 0.9386 & 0.3929 \\
 & LoRA          & \textbf{0.8686} & 0.9318 & 0.4927 & 0.3529 & 0.5180 & 0.6088 & 0.8413 & 0.4787 & 0.6499 & 0.6381 & 0.9242 & 0.0486 \\
 & FAPM          & 0.8609 & 0.9393 & 0.4953 & 0.2288 & 0.3640 & 0.6251 & 0.8469 & 0.4760 & 0.6646 & 0.6112 & 0.8484 & 0.0895 \\
 & DG-Hard (Ours)& 0.7517 & 0.9401 & \underline{0.8793} & 0.4113 & 0.5900 & \underline{0.7766} & 0.8425 & 0.4760 & \underline{0.7417} & \underline{0.7121} & 0.9314 & 0.6447 \\
\bottomrule
\end{tabular}%
}
\end{table}

\begin{table}[!htbp]
\centering
\caption{Per-benchmark held-out scores for Llama-3.2-3B-Instruct across
all (task, method) cells. \textbf{Bold} = best per column within the
task block; \underline{underline} = second-best (gap $\geq 0.001$ from
best). Avg.\ = mean of the nine held-out benchmarks. Results = on-task
\texttt{task\_\{name\}} score. Combined $= \mathrm{HM}(\overline{\text{\%
healed}},\, \overline{\text{\% preserved}})$ on the partitioned
held-out set, with per-benchmark \% healed and \% preserved clipped to
$[0, 100]$ before averaging. Pre-trained and Full-SFT have no defined
Combined. Bold/underline on Combined follow the same convention as
other columns, restricted to the seven repair methods (excluding the
two reference rows). ``Repair methods'' here covers both training-time
interventions (L1-reg, V-SoftMask~\citep{ke2023continual}, CoFi-Tune~\citep{zhang2024cofitune}, LoRA~\citep{hu2022lora}) and post-hoc
methods (WiSE-FT~\citep{wortsman2022robust}, FAPM~\citep{huang2025fapm}, DG-Hard); see \S\ref{sec:related-work} for
the full bucketing.}
\label{tab:bench-llama}
\resizebox{\textwidth}{!}{%
\footnotesize
\begin{tabular}{llcccccccccccc}
\toprule
Task & Method & ARC & GSM8K & HSwag & IFEval & Math & MMLU & MNLI & Trivia & TruthQA & Avg. & Results & Combined \\
\midrule
\multirow{9}{*}{MedQA}
 & Pre-trained   & \underline{0.7543} & 0.7218 & 0.6193 & \textbf{0.7784} & 0.4200 & 0.3474 & \underline{0.5025} & 0.3360 & 0.5606 & 0.5600 & 0.4949 & --- \\
 & Full-SFT      & 0.7440 & 0.7058 & \underline{0.6980} & 0.7599 & 0.4180 & \underline{0.5655} & 0.4665 & \textbf{0.5253} & \underline{0.6157} & \underline{0.6110} & \underline{0.5774} & --- \\
 & L1-reg        & 0.7457 & 0.7225 & 0.6787 & 0.7722 & 0.3980 & 0.4094 & 0.4745 & 0.4367 & 0.5973 & 0.5817 & 0.5695 & \underline{0.3181} \\
 & WiSE-FT       & 0.7517 & 0.7202 & 0.6727 & 0.7667 & 0.4120 & 0.3860 & 0.4710 & 0.4267 & 0.5961 & 0.5781 & \textbf{0.5829} & 0.1996 \\
 & V-SoftMask    & 0.7474 & 0.7096 & \textbf{0.7007} & 0.7557 & \underline{0.4360} & \textbf{0.5719} & 0.4610 & 0.5247 & 0.6108 & \textbf{0.6131} & 0.5742 & 0.0000 \\
 & CoFi-Tune     & 0.7534 & 0.7134 & 0.6460 & 0.7740 & \textbf{0.4620} & 0.3585 & 0.4415 & 0.3900 & 0.6059 & 0.5716 & 0.5687 & 0.0000 \\
 & LoRA          & 0.7483 & \textbf{0.7339} & 0.6440 & 0.7612 & 0.4300 & 0.3953 & 0.4740 & 0.4027 & \textbf{0.6193} & 0.5788 & 0.5734 & 0.2890 \\
 & FAPM          & \textbf{0.7594} & 0.7218 & 0.6180 & \underline{0.7766} & 0.4220 & 0.3538 & 0.5005 & 0.3427 & 0.5545 & 0.5610 & 0.5035 & 0.0318 \\
 & DG-Hard (Ours)& 0.7449 & \underline{0.7301} & 0.6693 & 0.7666 & 0.4280 & 0.3848 & \textbf{0.5045} & 0.4113 & 0.5924 & 0.5813 & 0.5774 & \textbf{0.6166} \\
\midrule
\multirow{9}{*}{WikiQA}
 & Pre-trained   & \underline{0.7543} & 0.7218 & \textbf{0.6193} & \underline{0.7784} & 0.4200 & 0.3474 & 0.5025 & 0.3360 & 0.5606 & 0.5600 & 0.9526 & --- \\
 & Full-SFT      & 0.7474 & 0.6520 & 0.5947 & 0.7354 & 0.4320 & 0.3404 & \underline{0.5580} & \textbf{0.4933} & \textbf{0.6108} & \underline{0.5738} & \textbf{0.9664} & --- \\
 & L1-reg        & 0.7389 & 0.7210 & 0.5347 & 0.7703 & 0.4200 & 0.3404 & 0.4920 & 0.4320 & 0.5655 & 0.5572 & 0.9638 & 0.3738 \\
 & WiSE-FT       & 0.7423 & 0.6808 & 0.5947 & 0.7535 & 0.4340 & \underline{0.3491} & 0.5425 & 0.4547 & 0.5936 & 0.5717 & 0.9629 & 0.5251 \\
 & V-SoftMask    & 0.7381 & 0.6368 & 0.5920 & 0.7423 & \textbf{0.4380} & 0.3392 & \textbf{0.5600} & \underline{0.4900} & 0.6083 & 0.5716 & 0.9663 & 0.1475 \\
 & CoFi-Tune     & 0.7193 & 0.6854 & 0.5687 & \textbf{0.7825} & 0.4200 & 0.3058 & 0.4790 & 0.3813 & \underline{0.6095} & 0.5502 & 0.9658 & \underline{0.5366} \\
 & LoRA          & 0.7329 & 0.7180 & 0.5687 & 0.7644 & \underline{0.4360} & 0.3450 & 0.4705 & 0.4380 & 0.5802 & 0.5615 & 0.9646 & 0.4850 \\
 & FAPM          & \textbf{0.7688} & \underline{0.7225} & \underline{0.6127} & 0.7743 & 0.4320 & \textbf{0.3532} & 0.5050 & 0.3407 & 0.5557 & 0.5628 & 0.9531 & 0.0485 \\
 & DG-Hard (Ours)& 0.7415 & \textbf{0.7293} & 0.5833 & 0.7720 & 0.4220 & 0.3433 & 0.5535 & 0.4833 & 0.5961 & \textbf{0.5805} & 0.9580 & \textbf{0.8881} \\
\midrule
\multirow{9}{*}{Winogrande}
 & Pre-trained   & 0.7543 & 0.7218 & 0.6193 & 0.7784 & 0.4200 & 0.3474 & 0.5025 & 0.3360 & \underline{0.5606} & 0.5600 & 0.5209 & --- \\
 & Full-SFT      & \underline{0.7765} & 0.6960 & \textbf{0.6673} & 0.7647 & 0.4140 & \underline{0.4275} & 0.5370 & 0.4533 & 0.4835 & 0.5800 & \textbf{0.8437} & --- \\
 & L1-reg        & 0.7167 & 0.7134 & 0.5947 & 0.7524 & 0.4120 & 0.3222 & 0.5150 & 0.4487 & 0.5422 & 0.5575 & 0.7956 & 0.4612 \\
 & WiSE-FT       & 0.7671 & 0.6967 & 0.6547 & 0.7715 & \textbf{0.4320} & 0.3731 & 0.5375 & \underline{0.4627} & 0.5483 & \underline{0.5826} & 0.7719 & \underline{0.8010} \\
 & V-SoftMask    & \textbf{0.7782} & 0.7058 & 0.6673 & 0.7572 & 0.4100 & \textbf{0.4351} & 0.5365 & 0.4507 & 0.4786 & 0.5799 & \underline{0.8421} & 0.0000 \\
 & CoFi-Tune     & 0.7218 & 0.6960 & 0.6193 & 0.7737 & 0.4320 & 0.3427 & \textbf{0.5765} & 0.3920 & \textbf{0.5692} & 0.5692 & 0.7301 & 0.5394 \\
 & LoRA          & 0.7560 & 0.6922 & 0.6173 & \textbf{0.7874} & 0.4320 & 0.3719 & 0.4900 & 0.3707 & 0.4933 & 0.5567 & 0.8114 & 0.1377 \\
 & FAPM          & 0.7585 & \underline{0.7233} & 0.6160 & \underline{0.7800} & 0.4240 & 0.3515 & 0.5035 & 0.3393 & 0.5557 & 0.5613 & 0.5178 & 0.0527 \\
 & DG-Hard (Ours)& 0.7517 & \textbf{0.7339} & 0.6580 & 0.7713 & 0.4060 & 0.3854 & \underline{0.5475} & \textbf{0.4680} & 0.5447 & \textbf{0.5852} & 0.8011 & \textbf{0.8066} \\
\midrule
\multirow{9}{*}{StrategyQA}
 & Pre-trained   & 0.7543 & \underline{0.7218} & \underline{0.6193} & \textbf{0.7784} & 0.4200 & 0.3474 & 0.5025 & 0.3360 & 0.5606 & 0.5600 & 0.6521 & --- \\
 & Full-SFT      & 0.7457 & 0.5519 & \textbf{0.6220} & 0.6854 & 0.3720 & \textbf{0.5228} & 0.5235 & \textbf{0.5427} & 0.5936 & 0.5733 & \underline{0.7220} & --- \\
 & L1-reg        & 0.7398 & 0.6740 & 0.6040 & 0.7228 & 0.4000 & 0.4766 & 0.5130 & 0.5213 & 0.6010 & 0.5836 & \textbf{0.7234} & 0.6898 \\
 & WiSE-FT       & 0.7363 & 0.6998 & 0.6100 & 0.7334 & 0.4380 & 0.4211 & 0.5190 & 0.4940 & \textbf{0.6083} & \textbf{0.5844} & 0.7074 & \textbf{0.7604} \\
 & V-SoftMask    & 0.7526 & 0.5557 & 0.6173 & 0.6808 & 0.3960 & \underline{0.5187} & \textbf{0.5320} & \underline{0.5393} & 0.5924 & 0.5761 & 0.7176 & 0.2954 \\
 & CoFi-Tune     & 0.7227 & 0.6884 & 0.5753 & 0.7708 & 0.4200 & 0.3491 & 0.5165 & 0.4427 & \underline{0.6059} & 0.5657 & 0.6929 & 0.6519 \\
 & LoRA          & 0.7389 & \textbf{0.7240} & 0.6020 & \underline{0.7730} & \textbf{0.4400} & 0.3164 & 0.5070 & 0.3687 & 0.5594 & 0.5588 & 0.6623 & 0.1000 \\
 & FAPM          & \textbf{0.7551} & 0.7187 & 0.6153 & 0.7729 & 0.4160 & 0.3485 & 0.5050 & 0.3367 & 0.5508 & 0.5577 & 0.6463 & 0.0066 \\
 & DG-Hard (Ours)& 0.7270 & 0.7058 & 0.6080 & 0.7324 & 0.4400 & 0.3772 & \underline{0.5250} & 0.4767 & 0.5973 & 0.5766 & 0.6914 & \underline{0.6980} \\
\midrule
\multirow{9}{*}{BoolQ}
 & Pre-trained   & 0.7543 & 0.7218 & \underline{0.6193} & \textbf{0.7784} & 0.4200 & 0.3474 & \underline{0.5025} & 0.3360 & 0.5606 & 0.5600 & 0.7859 & --- \\
 & Full-SFT      & \textbf{0.7628} & 0.6876 & \textbf{0.6353} & 0.7187 & 0.4040 & \textbf{0.4801} & 0.4900 & \textbf{0.4967} & 0.5679 & \textbf{0.5826} & \underline{0.8902} & --- \\
 & L1-reg        & 0.7543 & 0.7233 & 0.5520 & \underline{0.7756} & 0.4160 & 0.3661 & 0.4845 & 0.4807 & \underline{0.5887} & 0.5712 & 0.8872 & \textbf{0.6793} \\
 & WiSE-FT       & \underline{0.7551} & 0.7240 & 0.6120 & 0.7690 & 0.4240 & 0.3813 & 0.4995 & 0.4647 & 0.5875 & 0.5797 & 0.8731 & \underline{0.6714} \\
 & V-SoftMask    & 0.7517 & 0.6657 & 0.6067 & 0.7046 & \textbf{0.4440} & \underline{0.4520} & 0.4715 & \underline{0.4847} & 0.5875 & 0.5743 & \textbf{0.8951} & 0.0000 \\
 & CoFi-Tune     & 0.7346 & 0.7119 & 0.5687 & 0.7681 & 0.4340 & 0.3105 & 0.4555 & 0.4313 & \textbf{0.5985} & 0.5570 & 0.8832 & 0.4282 \\
 & LoRA          & 0.7423 & \textbf{0.7377} & 0.5607 & 0.7606 & \underline{0.4400} & 0.3298 & 0.4685 & 0.4567 & 0.5789 & 0.5639 & 0.8743 & 0.5211 \\
 & FAPM          & 0.7551 & 0.7149 & 0.6167 & 0.7693 & 0.4140 & 0.3538 & 0.5000 & 0.3387 & 0.5532 & 0.5573 & 0.7856 & 0.0626 \\
 & DG-Hard (Ours)& 0.7526 & \underline{0.7286} & 0.6147 & 0.7686 & 0.4280 & 0.3673 & \textbf{0.5065} & 0.4720 & 0.5802 & \underline{0.5798} & 0.8633 & 0.6457 \\
\midrule
\multirow{9}{*}{ReClor}
 & Pre-trained   & 0.7543 & 0.7218 & 0.6193 & \underline{0.7784} & 0.4200 & 0.3474 & 0.5025 & 0.3360 & 0.5606 & 0.5600 & 0.4860 & --- \\
 & Full-SFT      & \underline{0.7628} & \underline{0.7468} & \textbf{0.6573} & 0.7558 & 0.4300 & \textbf{0.4386} & \underline{0.5290} & \textbf{0.4860} & \textbf{0.6255} & \textbf{0.6035} & \textbf{0.6800} & --- \\
 & L1-reg        & 0.7526 & 0.7354 & 0.6420 & 0.7548 & \underline{0.4460} & 0.3731 & \textbf{0.5330} & 0.4427 & 0.6181 & 0.5886 & 0.6760 & \underline{0.7648} \\
 & WiSE-FT       & \textbf{0.7688} & 0.7263 & 0.6373 & \textbf{0.7856} & \textbf{0.4500} & 0.3532 & 0.5155 & 0.4467 & 0.6083 & 0.5880 & 0.6180 & 0.6692 \\
 & V-SoftMask    & 0.7628 & \textbf{0.7521} & 0.6573 & 0.7492 & 0.4320 & \underline{0.4310} & 0.5260 & \underline{0.4780} & \underline{0.6230} & \underline{0.6013} & 0.6800 & \textbf{0.9777} \\
 & CoFi-Tune     & 0.7355 & 0.7233 & 0.6053 & 0.7689 & 0.4360 & 0.3310 & 0.5260 & 0.3907 & 0.6157 & 0.5703 & 0.5960 & 0.4655 \\
 & LoRA          & 0.7415 & 0.7233 & 0.6140 & 0.7764 & 0.4420 & 0.3257 & 0.5155 & 0.3640 & 0.5887 & 0.5657 & 0.5620 & 0.2686 \\
 & FAPM          & 0.7560 & 0.7187 & 0.6187 & 0.7740 & 0.4240 & 0.3596 & 0.5030 & 0.3453 & 0.5508 & 0.5611 & 0.4900 & 0.0938 \\
 & DG-Hard (Ours)& 0.7560 & 0.7324 & 0.6340 & 0.7733 & 0.4420 & 0.3386 & 0.5255 & 0.4567 & 0.6059 & 0.5849 & 0.6040 & 0.6414 \\
\midrule
\multirow{9}{*}{RTE}
 & Pre-trained   & \underline{0.7543} & \underline{0.7218} & \textbf{0.6193} & \underline{0.7784} & 0.4200 & 0.3474 & 0.5025 & 0.3360 & 0.5606 & 0.5600 & 0.5957 & --- \\
 & Full-SFT      & 0.7346 & 0.6308 & 0.6013 & 0.7031 & 0.3660 & \textbf{0.5228} & 0.5870 & \textbf{0.5053} & 0.5802 & 0.5812 & \textbf{0.9061} & --- \\
 & L1-reg        & 0.7082 & 0.6960 & 0.5847 & 0.7508 & 0.3980 & 0.4398 & \underline{0.5895} & 0.4693 & \textbf{0.6034} & \underline{0.5822} & 0.8845 & \textbf{0.7042} \\
 & WiSE-FT       & 0.7142 & 0.7051 & 0.5853 & 0.7686 & 0.4180 & 0.3889 & 0.5630 & 0.4620 & 0.6010 & 0.5784 & 0.8773 & \underline{0.6895} \\
 & V-SoftMask    & 0.7321 & 0.6240 & 0.6087 & 0.7003 & 0.3800 & \underline{0.5187} & \textbf{0.6050} & 0.5053 & 0.5924 & \textbf{0.5852} & 0.9061 & 0.1590 \\
 & CoFi-Tune     & 0.7014 & 0.6808 & 0.5600 & 0.7534 & \underline{0.4280} & 0.3205 & 0.5225 & 0.4160 & 0.6010 & 0.5537 & 0.8736 & 0.3582 \\
 & LoRA          & 0.7321 & \textbf{0.7309} & 0.6053 & \textbf{0.7860} & 0.4220 & 0.3392 & 0.5320 & 0.3727 & 0.5753 & 0.5662 & 0.8448 & 0.3173 \\
 & FAPM          & \textbf{0.7568} & 0.7165 & \underline{0.6127} & 0.7750 & 0.4080 & 0.3520 & 0.5045 & 0.3407 & 0.5618 & 0.5587 & 0.5993 & 0.0505 \\
 & DG-Hard (Ours)& 0.7201 & 0.7081 & 0.5840 & 0.7726 & \textbf{0.4300} & 0.3667 & 0.5620 & 0.4467 & 0.6034 & 0.5771 & 0.8592 & 0.6398 \\
\bottomrule
\end{tabular}%
}
\end{table}

\subsection{Method-specific configurations}
\label{app:method-configs}

\textbf{L1-reg} adds an L1 penalty
$\|W - W_{\mathrm{base}}\|_1$ to the loss with
$\lambda = 1.0 \times 10^{-6}$.
\textbf{LoRA} uses rank $r = 16$, $\alpha = 32$, dropout $0.05$, and
auto-selected target modules.
\textbf{WiSE-FT} sets the mixing weight $\alpha = 0.5$, applied
post-hoc on the Full-SFT checkpoint to give the linear weight average
$W^{*} = \tfrac{1}{2}W_{\mathrm{base}} + \tfrac{1}{2}W_{\mathrm{ft}}$,
the canonical setting from \citet{wortsman2022robust}.
\textbf{FAPM} reverts the entries of $\Delta$ ranked lowest under the
forgetting-aware FAPM score, which
combines absolute change with a relative-change penalty against
$W_{\mathrm{base}}$); we use the published $90\%$ reversion rate.
\textbf{V-SoftMask} (our reproduction of DAS,
\citealp{ke2023continual}) uses a calibration batch size of $4$ and
applies mask $= (1 - \text{normalised gradient importance})$, fixed
for the SFT run.
\textbf{CoFi-Tune} uses the two-stage configuration of
\citet{zhang2024cofitune} verbatim (full mechanism in
\S\ref{sec:related-work}); we run the coarse layer-range filter at
the second-quartile setting $(N \times 25\%, N \times 50\%]$.
\textbf{DG-Hard} applies the $\omega(\beta) \cdot \hat{\sigma}_{\text{eff}}$
hard threshold of \S\ref{sec:method}; it has no tuning knobs and uses
no calibration data.
\textbf{TIES}~\citep{yadav2023ties} keeps the top-$k$ entries of
$|\Delta|$ by magnitude, sign-elects, and averages; we use
\texttt{keep\_ratio}~$=0.2$ ($k=20\%$) with mixing weight $\lambda=1$.
\textbf{DARE}~\citep{yu2023language} drops entries of $\Delta$ with
probability $p$ and rescales the survivors by $1/(1-p)$; we use
\texttt{drop\_prob}~$=0.5$ with the rescaled (rather than raw) variant.
TIES and DARE are evaluated only in the App.~\ref{app:abl-merging}
ablation, not in the headline comparison of Tab.~\ref{tab:bench-qwen}
and Tab.~\ref{tab:bench-llama}.

\subsection{Inference engine and decoding}
\label{app:inference}

\textbf{All inference uses vLLM} with \texttt{--enforce-eager},
\texttt{max\_model\_len{=}8192}, \texttt{dtype{=}bfloat16}, greedy
decoding (temperature $= 0$), \emph{no} \texttt{max\_tokens} cap, and
the model's native chat template applied via the tokenizer's
\texttt{apply\_chat\_template} method. Greedy decoding under vLLM's
continuous batching is not bit-deterministic: the same prompt can
produce slightly different completions on re-runs, with a measured
\textbf{drift floor of about $\pm 1$ pp per benchmark}. We treat any
delta smaller than this floor as noise, and use a $\textbf{3 pp}$
\textbf{significance threshold} throughout the paper. The threshold
is conservative ($\approx 3\times$ the drift floor) and is chosen to
capture genuine but moderate fine-tuning effects while excluding
sampling drift. Output scoring strips
\texttt{<think>}\,\ldots\,\texttt{</think>} blocks and reasoning
preludes before metric computation.

\section{Other experiments and ablations}
\label{app:ablations}

This appendix collects three experiments supporting the design
choices of DG-Hard. App.~\ref{app:abl-noise-heatmap} quantifies which
weight classes carry the noise mass that DG-Hard removes
(layer-class noise concentration); App.~\ref{app:abl-mask} verifies
the layer-mask choice via a causal test; App.~\ref{app:abl-merging}
compares against element-wise merging baselines at canonical and
matched-rollback operating points.

\subsection{Layer-wise noise concentration}
\label{app:abl-noise-heatmap}

For each (model, task) fine-tune, we cache the SVD of every 2D weight
delta and classify each matrix by its layer class (e.g.,
\texttt{mlp.up\_proj}, \texttt{attn.q\_proj}). Per class we report the
mean noise-energy fraction (the share of singular-value-squared mass
sitting below the MP edge $\lambda_{\mathrm{MP}}$ of
\eqref{eq:mp-edge}) and the total norm-share (the fraction of the
network's Frobenius$^{2}$ delta mass attributable to the class). Their
product is the operationally meaningful noise mass per class under the
signal-plus-noise model of \eqref{eq:s+n}. Averaged across six
(model, task) pairs, the MLP gate and up projections concentrate the
bulk of the noise mass that DG-Hard removes
(Tab.~\ref{tab:abl-noise-class}); attention and embedding classes
contribute substantially less.

\begin{table}[h]
\centering
\small
\setlength{\tabcolsep}{6pt}
\caption{Per-class noise ranking, averaged over six (model, task)
pairs. Mean noise is the fraction of squared singular-value mass below
$\lambda_{\mathrm{MP}}$. Mean norm-share is the class share of the
network's total Frobenius$^{2}$ delta. Their product is the noise mass
per class. Top-two rows in bold.}
\label{tab:abl-noise-class}
\begin{tabular}{lccc}
\toprule
Class & Mean noise & Mean norm-share & Noise $\times$ norm-share \\
\midrule
\textbf{\texttt{mlp.up\_proj}}   & \textbf{0.93} & \textbf{0.231} & \textbf{0.213} \\
\textbf{\texttt{mlp.gate\_proj}} & \textbf{0.93} & \textbf{0.214} & \textbf{0.196} \\
\texttt{mlp.down\_proj}          & 0.55 & 0.238 & 0.131 \\
\texttt{attn.v\_proj}            & 0.41 & 0.045 & 0.018 \\
\texttt{attn.o\_proj}            & 0.10 & 0.099 & 0.010 \\
\texttt{attn.k\_proj}            & 0.37 & 0.022 & 0.008 \\
\texttt{embed\_tokens}           & $\approx$1.00 & 0.026 & 0.026 \\
\bottomrule
\end{tabular}
\end{table}

\paragraph{Metric note.} The mean-noise column above (energy fraction
of squared singular values below $\lambda_{\mathrm{MP}}$) and the
MP-fit score of \eqref{eq:mp-fit} (a Kolmogorov-Smirnov distance
between the empirical and theoretical CDFs) measure different things:
the former is the share of squared mass under the bulk edge, the
latter is how closely the spectrum's shape matches the MP density. The
numerical proximity of the \texttt{mlp.up\_proj} row's $0.93$ to the
$93.1\%$ MP-fit reported in App.~\ref{app:low-rank-evidence} is
therefore coincidental, not the same quantity. The
\texttt{embed\_tokens} value $\approx 1.00$ is a degenerate case in
which essentially all of the FT delta's spectral mass sits under the
bulk edge: there is no task-specific signal in the embedding update on
these (model, task) pairs, consistent with
App.~\ref{app:low-rank-evidence}'s claim that the learned signal is
concentrated in a small number of spike directions, just not in the
embedding for these fine-tunes.

The aggregate ranking conceals a sharp cross-family asymmetry in
attention noise (Tab.~\ref{tab:abl-noise-family}): on Llama-3.2-3B,
\texttt{attn.q\_proj} has mean below-edge mass $0.025$, while on
Qwen3.5-4B the same class shows $0.91$, a $36\times$ ratio. The
\texttt{attn.o\_proj} class shows a $7\times$ ratio in the same
direction. MLP classes are family-invariant (ratios within $1.16$).

\begin{table}[h]
\centering
\small
\setlength{\tabcolsep}{6pt}
\caption{Cross-family asymmetry in mean noise-energy fraction.
Attention classes diverge sharply between Llama-3.2-3B-Instruct and
Qwen3.5-4B; MLP classes do not.}
\label{tab:abl-noise-family}
\begin{tabular}{lccc}
\toprule
Class & Llama mean noise & Qwen mean noise & Ratio \\
\midrule
\texttt{mlp.gate\_proj} & 0.93 & 0.94 & 1.01 \\
\texttt{mlp.up\_proj}   & 0.92 & 0.93 & 1.01 \\
\texttt{mlp.down\_proj} & 0.51 & 0.59 & 1.16 \\
\texttt{attn.q\_proj}   & 0.025 & 0.91 & $36\times$ \\
\texttt{attn.o\_proj}   & 0.027 & 0.19 & $7\times$ \\
\texttt{attn.k\_proj}   & 0.36 & 0.39 & 1.08 \\
\texttt{attn.v\_proj}   & 0.44 & 0.35 & 0.80 \\
\bottomrule
\end{tabular}
\end{table}

The family asymmetry is consistent with the hypothesis that reasoning
post-training (long chain-of-thought RL or distillation, present in
Qwen3.5-4B but not in Llama-3.2-3B-Instruct) disperses attention
weights and leaves more below-edge mass in the FT delta, but this
explanation is preliminary and untested in the present run. Regardless
of the underlying cause, the gate and up MLP classes carry the
dominant share of removable noise mass in both families, motivating
the gate$+$up layer mask whose causal effect is tested in
\S\ref{app:abl-mask}.

\subsection{Layer-mask causal test}
\label{app:abl-mask}

App.~\ref{app:abl-noise-heatmap} ranks 2D weight classes by noise mass
and identifies \texttt{gate\_proj} and \texttt{up\_proj} as the
top-two carriers on every (model, task) pair. This appendix verifies
that ranking causally: we vary the set of layers pruned by DG-Hard and
measure cross-domain accuracy. The protocol is DG-Hard canonical
(scale $= 1.0$, DG $\sigma$) on two pairs, Qwen3.5-4B with Winogrande
(severe FT damage) and Llama-3.2-3B with MedQA (mild FT damage), under
four masks: \texttt{ALL} (every 2D matrix), \texttt{mlp\_only}
(\texttt{\textbackslash.mlp\textbackslash.}), \texttt{attn\_only}
(\texttt{\textbackslash.self\_attn\textbackslash.}), and
\texttt{gate\_up} (\texttt{\textbackslash.mlp\textbackslash.(gate$|$up)\_proj}).
Cells outside the active mask pass through with the full FT delta (no
shrinkage); we report Frobenius retention $r$ alongside benchmark
accuracies and the protocol-aligned Combined score. The four masks
above target attention/MLP partitions of the decoder; the
\texttt{embed\_tokens} and \texttt{lm\_head} matrices, which are in
scope for the headline DG-Hard configuration
(App.~\ref{app:repair-scope}), are kept at their \texttt{ALL}-mask
repaired state in all four conditions of this ablation. The focus of
this experiment is the attention-vs-MLP partition; the embedding
contribution is characterised separately in
App.~\ref{app:abl-noise-heatmap}.

\paragraph{Column conventions.} \texttt{task\_win} and
\texttt{task\_med} are on-task accuracy on Winogrande and MedQA,
respectively. The Combined column applies the partition-conditional
harmonic mean of \eqref{eq:combined-def}; when $U = \emptyset$
(Tab.~\ref{tab:abl-mask-qwen}) the non-damage term of
\eqref{eq:metric-defs} is undefined ($0/0$) and we adopt the
convention $\text{non-damage} = 100$, matching the implementation.
With this convention the \texttt{attn\_only} row of
Tab.~\ref{tab:abl-mask-qwen} collapses to Combined~$=0.0$ because
percent-healed is negative on the damaged benches (GSM8K, ARC-C),
which the harmonic mean drives to zero independently of the U-term.

\begin{table}[h]
\centering
\small
\setlength{\tabcolsep}{4pt}
\caption{Mask ablation on Qwen3.5-4B + Winogrande (severe FT damage).
$D = \{\text{arc\_challenge}, \text{gsm8k}\}$,
$I = \{\text{hellaswag}, \text{ifeval}, \text{mmlu}\}$,
$U = \emptyset$. Combined uses the protocol harmonic mean over the
partition. Winning mask on Combined in bold.}
\label{tab:abl-mask-qwen}
\resizebox{\linewidth}{!}{%
\begin{tabular}{lcccccccc}
\toprule
Mask & GSM8K & ARC-C & HSwag & IFEval & MMLU & task\_win & $r$ & Combined \\
\midrule
ALL        & 87.7 & 73.9 & 85.0 & 36.1 & 75.7 & 89.4 & 0.314 & 72.6 \\
mlp\_only  & 89.2 & 72.6 & 86.7 & 35.7 & 78.2 & 89.0 & 0.633 & \textbf{73.7} \\
attn\_only & 58.5 & 55.3 & 87.2 & 41.2 & 75.0 & 89.7 & 0.966 & 0.0 \\
gate\_up   & 87.6 & 67.7 & 87.0 & 39.4 & 77.9 & 89.6 & 0.780 & 67.4 \\
\bottomrule
\end{tabular}}
\end{table}

\begin{table}[h]
\centering
\small
\setlength{\tabcolsep}{4pt}
\caption{Mask ablation on Llama-3.2-3B + MedQA (mild FT damage).
$D = \{\text{gsm8k}\}$, $I = \{\text{hellaswag}, \text{mmlu}\}$,
$U = \{\text{arc\_challenge}, \text{ifeval}\}$. Winning mask on
Combined in bold.}
\label{tab:abl-mask-llama}
\resizebox{\linewidth}{!}{%
\begin{tabular}{lcccccccc}
\toprule
Mask & GSM8K & ARC-C & HSwag & IFEval & MMLU & task\_med & $r$ & Combined \\
\midrule
ALL        & 73.8 & 74.6 & 67.1 & 75.9 & 38.7 & 57.7 & 0.186 & 79.5 \\
mlp\_only  & 71.0 & 74.3 & 68.1 & 75.4 & 43.0 & 57.3 & 0.540 & 79.5 \\
attn\_only & 70.4 & 73.9 & 69.6 & 75.9 & 49.2 & 57.7 & 0.874 & \textbf{82.9} \\
gate\_up   & 69.7 & 74.4 & 69.5 & 76.9 & 49.1 & 57.5 & 0.730 & 75.3 \\
\bottomrule
\end{tabular}}
\end{table}

\paragraph{Pattern A (Qwen, severe FT).}
\texttt{ALL}, \texttt{mlp\_only}, and \texttt{gate\_up} cluster
closely on the per-benchmark damaged accuracies (GSM8K 87.6--89.2,
ARC-C 67.7--73.9); \texttt{attn\_only} collapses on the damaged
benches with GSM8K dropping to 58.5 (vs.\ 87.7 under \texttt{ALL}) and
ARC-C to 55.3 (vs.\ 73.9). The protocol-aligned Combined column is
more discriminating still: \texttt{attn\_only} scores 0.0 because it
has negative percent healed on the damaged benches, which the
harmonic mean penalises. Repairing only attention while leaving the
MLP delta unshrunken keeps too much noise in the model, regardless of
how aggressively attention is repaired. Repairing the MLP class is
necessary; repairing only attention is insufficient.

\paragraph{Pattern B (Llama, mild FT).}
All four masks cluster tightly across the per-benchmark accuracies,
and Combined separates them only mildly. With a mild FT delta, the
absolute amount of noise to remove is small, so any reasonable repair
yields similar cross-domain accuracy. We treat this as a
\emph{negative} result: the variance is genuinely small, not evidence
that all masks are equal. The cross-family contrast is informative.
On Llama, attention is intrinsically low-noise
(\S\ref{app:abl-noise-heatmap}), so \texttt{attn\_only} acts almost as
a pass-through and lands near \texttt{ALL}. On Qwen,
\texttt{attn.q\_proj} carries noise-mass 0.91 (vs.\ 0.025 on Llama),
so leaving the MLP unrepaired collapses the cell. \texttt{attn\_only}
is therefore not a fair control across families: it isolates a
different noise source on each.

Across both pairs, \texttt{gate\_up} is the highest-$r$ mask whose
per-benchmark accuracies match \texttt{ALL} (retention 0.78 and 0.73
vs.\ 0.31 and 0.19), capturing the top-two noise classes from
\S\ref{app:abl-noise-heatmap} while preserving 73 to 78 percent of the
FT delta: the operational sweet spot of maximum FT preservation with
minimum noise leakage.

\subsection{Comparison with element-wise merging baselines}
\label{app:abl-merging}

We compare DG-Hard against three element-wise merging baselines on
two pairs (Qwen3.5-4B + Winogrande, severe FT, and
Llama-3.2-3B + MedQA, mild FT): WiSE-FT and Task Arithmetic
\citep{wortsman2022robust,ilharco2023editing}, which are
mathematically identical for a single (base, FT) pair and which we
report jointly; TIES \citep{yadav2023ties}, which keeps the top
$k$ entries of $|\Delta|$, sign-elects, and averages; and DARE
\citep{yu2023language}, which randomly drops entries with probability
$p$ and rescales surviving entries by $1/(1-p)$. We use ``DARE'' to
denote the canonical rescaled form (\texttt{rescale=True}) and
``DARE-Linear'' to denote the variant without rescaling
(\texttt{rescale=False}). Each method is evaluated at two operating
points. Part A uses each baseline at its published default
(WiSE-FT/Task-Arith $\alpha{=}0.5$; TIES \texttt{keep\_ratio}$=0.2$
with $\lambda{=}1$; DARE \texttt{drop\_prob}$=0.5$ with
\texttt{rescale=True}; DG-Hard at its parameter-free RMT-optimal
threshold).
Part B bisects every baseline to the same Frobenius rollback
$r{=}0.5$, isolating shape effects from magnitude effects; this
matched-rollback comparison is the central methodological control.
For DARE in Part B we use the linear no-rescale form (DARE-Linear),
since DARE with \texttt{rescale=True} at $p{=}0.5$ has expected
Frobenius ratio $\sqrt{1/(1-p)} \approx 1.41$, which is incompatible
with the matched $r{=}0.5$ target.

\begin{table}[h]
\centering
\footnotesize
\setlength{\tabcolsep}{4pt}
\caption{Merging baselines on Qwen3.5-4B + Winogrande (severe FT
damage). $D = \{\text{arc\_challenge}, \text{gsm8k}\}$,
$I = \{\text{hellaswag}, \text{ifeval}, \text{mmlu}\}$,
$U = \emptyset$. Combined uses the protocol harmonic mean over the
partition. Winning Combined per part in bold. DARE's~\citep{yu2023language} expected
Frobenius ratio is $\sqrt{1/(1-p)} \approx 1.41$ at $p{=}0.5$; the
realized $r{=}1.155$ reflects single-seed mask variance. Combined of
$0$ indicates a sub-statistic clipped to its floor (e.g., \% preserved
non-positive on the improved partition), which collapses the harmonic
mean.}
\label{tab:abl-merge-qwen}
\resizebox{\linewidth}{!}{%
\begin{tabular}{lcccccccc}
\toprule
Method & $r$ & GSM8K & ARC-C & HSwag & IFEval & MMLU & task\_win & Combined \\
\midrule
\multicolumn{9}{l}{\emph{Part A: canonical operating points}} \\
\midrule
WiSE-FT / Task-Arith $\alpha{=}0.5$ & 0.500 & 91.1 & 81.7 & 60.9 & 35.5 & 72.6 & 85.7 & 67.5 \\
TIES canonical                      & 1.000 & 57.3 & 62.6 & 87.1 & 41.4 & 75.8 & 89.8 & 0.0 \\
DARE canonical                      & 1.155 & 56.3 & 65.8 & 87.0 & 43.2 & 75.3 & 89.6 & 7.9 \\
DG-Hard canonical                   & 0.314 & 87.7 & 73.9 & 85.0 & 36.1 & 75.7 & 89.4 & \textbf{72.6} \\
\midrule
\multicolumn{9}{l}{\emph{Part B: matched Frobenius rollback $r{=}0.5$}} \\
\midrule
WiSE-FT / Task-Arith $r{=}0.5$ & 0.500 & 91.1 & 81.7 & 60.9 & 35.5 & 72.6 & 85.7 & 67.5 \\
TIES $r{=}0.5$                 & 0.504 & 92.7 & 87.4 & 47.3 & 35.2 & 65.0 & 68.0 & 21.3 \\
DARE-Linear $r{=}0.5$          & 0.354 & 92.4 & 87.1 & 47.3 & 35.4 & 64.8 & 79.7 & 22.0 \\
DG-Hard $r{=}0.5$              & 0.500 & 90.3 & 75.3 & 87.1 & 38.3 & 79.8 & 89.3 & \textbf{81.5} \\
\bottomrule
\end{tabular}}
\end{table}

\begin{table}[h]
\centering
\footnotesize
\setlength{\tabcolsep}{4pt}
\caption{Merging baselines on Llama-3.2-3B + MedQA (mild FT damage).
$D = \{\text{gsm8k}\}$, $I = \{\text{hellaswag}, \text{mmlu}\}$,
$U = \{\text{arc\_challenge}, \text{ifeval}\}$. Winning Combined per
part in bold. DARE's~\citep{yu2023language} expected Frobenius ratio is
$\sqrt{1/(1-p)} \approx 1.41$ at $p{=}0.5$; the realized $r{=}1.155$
reflects single-seed mask variance. Combined of $0$ indicates a
sub-statistic clipped to its floor (e.g., \% preserved non-positive on
the improved partition), which collapses the harmonic mean.}
\label{tab:abl-merge-llama}
\resizebox{\linewidth}{!}{%
\begin{tabular}{lcccccccc}
\toprule
Method & $r$ & GSM8K & ARC-C & HSwag & IFEval & MMLU & task\_med & Combined \\
\midrule
\multicolumn{9}{l}{\emph{Part A: canonical operating points}} \\
\midrule
WiSE-FT / Task-Arith $\alpha{=}0.5$ & 0.501 & 71.1 & 74.6 & 67.5 & 76.9 & 38.1 & 58.7 & 72.8 \\
TIES canonical                      & 1.000 & 67.2 & 74.5 & 70.1 & 75.2 & 55.7 & 58.1 & 5.4 \\
DARE canonical                      & 1.155 & 66.8 & 74.4 & 70.2 & 76.0 & 55.3 & 57.7 & 0.0 \\
DG-Hard canonical                   & 0.186 & 73.8 & 74.6 & 67.1 & 75.9 & 38.7 & 57.7 & \textbf{79.5} \\
\midrule
\multicolumn{9}{l}{\emph{Part B: matched Frobenius rollback $r{=}0.5$}} \\
\midrule
WiSE-FT / Task-Arith $r{=}0.5$ & 0.501 & 71.1 & 74.6 & 67.5 & 76.9 & 38.1 & 58.7 & 72.8 \\
TIES $r{=}0.5$                 & 0.503 & 70.4 & 74.5 & 64.8 & 78.3 & 34.0 & 57.6 & 47.4 \\
DARE-Linear $r{=}0.5$          & 0.354 & 70.1 & 74.8 & 65.3 & 78.0 & 31.3 & 58.0 & 42.5 \\
DG-Hard $r{=}0.5$              & 0.427 & 72.8 & 74.1 & 68.8 & 75.9 & 44.7 & 57.4 & \textbf{90.4} \\
\bottomrule
\end{tabular}}
\end{table}

\paragraph{Canonical operating points.}
On Qwen + Winogrande, DG-Hard wins on Combined (72.6 vs.\ 67.5 for
WiSE-FT/Task-Arith) and improves the damaged benches relative to
TIES/DARE (GSM8K 87.7 vs.\ 57.3/56.3; ARC-C 73.9 vs.\ 62.6/65.8). On
Llama + MedQA, TIES and DARE post slightly higher MMLU than DG-Hard
(55.7/55.3 vs.\ 38.7), yet their Combined collapses to 5.4 and 0.0
respectively. The explanation is visible in the $r$ column: at
canonical settings TIES and DARE sit at $r{\approx}1$, so they do
not shrink the FT delta and therefore heal a negligible fraction of
the FT-induced damage on GSM8K. The partition-conditional Combined
headline reveals this gap; an unweighted per-benchmark mean would
not. WiSE-FT and Task Arithmetic produce identical outputs for a
single (base, FT) pair, so we report a single row and credit both
citations.

\paragraph{Matched Frobenius rollback $r{=}0.5$.}
With magnitude held fixed, DG-Hard wins both pairs on Combined
($+14.0$~pp on Qwen + Winogrande, 81.5 vs.\ 67.5; $+17.6$~pp on
Llama + MedQA, 90.4 vs.\ 72.8). The per-benchmark breakdown shows
why: TIES at $r{=}0.5$ collapses HellaSwag (47.3 vs.\ 87.1 under
DG-Hard) and DARE-Linear at $r{=}0.5$ collapses ARC-C relative to
DG-Hard's distribution of mass while also collapsing HellaSwag
(47.3). DG-Hard is the only method that rolls back to
$r{\approx}0.5$ without driving any of the five benchmarks to a
degenerate value. TIES and DARE-Linear sparsify in the standard
parameter basis while DG-Hard sparsifies in the singular-vector
basis; both bets land at the same Frobenius rollback, but spectral
sparsity beats coordinate sparsity at the matched rollback point.



\end{document}